\documentclass[letterpaper,journal]{IEEEtran}

\usepackage{cite}
\usepackage{amsmath,amssymb,amsfonts}
\usepackage{algorithm}
\usepackage{algorithmic}
\usepackage{graphicx}
\usepackage{textcomp}
\usepackage{xcolor}
\def\BibTeX{{\rm B\kern-.05em{\sc i\kern-.025em b}\kern-.08em
    T\kern-.1667em\lower.7ex\hbox{E}\kern-.125emX}}
\usepackage{multirow}
\usepackage{amsfonts}
\usepackage{pifont}
\usepackage{listings}
\usepackage{txfonts}
\usepackage{graphics}
\usepackage{flafter}
\usepackage{indentfirst}
\usepackage{epsfig}
\usepackage{mathrsfs}
\usepackage{float}
\usepackage{longtable}
\usepackage{cite}
\usepackage{graphicx}
\usepackage{rotating}
\usepackage{diagbox}
\usepackage{tikz}
\usetikzlibrary{positioning}
\usepackage{tikz}
\usepackage{pdflscape}
\usepackage{bm}
\usepackage{float}
\usepackage[switch]{lineno}
\usepackage[T1]{fontenc}
\usepackage{mathrsfs}

\usepackage{subcaption}
\ifCLASSINFOpdf
\else
\fi

\begin{document}

\title{Granular-Balls based Fuzzy Twin Support Vector Machine for Classification}


\author{Lixi~Zhao, Weiping Ding, \emph{Senior Member, IEEE}, Duoqian Miao,~Guangming~Lang
\thanks{Lixi Zhao and Guangming Lang are with the School of Mathematics and Statistics, Changsha University of Science and Technology, Changsha, Hunan 410114, China; Weiping Ding is with the School of Information Science and Technology, Nantong University, Nantong, Jiangsu 226019, China; Duoqian Miao is with the Department of Computer Science and Technology, Tongji University, Shanghai, 201804, China (email: zhaolixi0915@163.com, langguangming1984@126.com, dwp9988@163.com, dqmiao@tongji.edu.cn; Corresponding author: Guangming Lang).}
}

\markboth{**}%
{Shell \MakeLowercase{\textit{et al.}}: Bare Demo of IEEEtran.cls for IEEE Journals}





%




\maketitle
\begin{abstract}
The twin support vector machine (TWSVM) classifier has attracted increasing attention because of its low computational complexity. However, its performance tends to degrade when samples are affected by noise. The granular-ball fuzzy support vector machine (GBFSVM) classifier partly alleviates the adverse effects of noise, but it relies solely on the distance between the granular-ball's center and the class center to design the granular-ball membership function.
In this paper, we first introduce the granular-ball twin support vector machine (GBTWSVM) classifier, which integrates granular-ball computing (GBC) with the twin support vector machine (TWSVM) classifier. By replacing traditional point inputs with granular-balls, we demonstrate how to derive a pair of non-parallel hyperplanes for the GBTWSVM classifier by solving a quadratic programming problem. Subsequently, we design the membership and non-membership functions of granular-balls using Pythagorean fuzzy sets to differentiate the contributions of granular-balls in various regions. Additionally, we develop the granular-ball fuzzy twin support vector machine (GBFTSVM) classifier by incorporating GBC with the fuzzy twin support vector machine (FTSVM) classifier. We demonstrate how to derive a pair of non-parallel hyperplanes for the GBFTSVM classifier by solving a quadratic programming problem. We also design algorithms for the GBTSVM classifier and the GBFTSVM classifier. Finally, the superior classification performance of the GBTWSVM classifier and the GBFTSVM classifier on 20 benchmark datasets underscores their scalability, efficiency, and robustness in tackling classification tasks.
\end{abstract}


\IEEEpeerreviewmaketitle

\section{Introduction}
\IEEEPARstart{S}{upport} Vector Machine (SVM), developed by Vapnik in 1995~\cite{V.N. Vapnik_1998}, is a machine learning technique for both classification and regression tasks. It is grounded in statistical learning theory and structural risk minimization theory. Up to now, SVM and its variants have shown unique advantages in analyzing high-dimensional and nonlinear datasets~\cite{B. Heisele_2001, O.L. Mangasarian_2006, N.Y. Deng_2012, V.N. Vapnik_2017}.
Especially, TWSVM generates two non-parallel hyperplanes instead of a single hyperplane for classification~\cite{Jayadeva_2007, R. Khemchandani_2009, Z. Qi_2013,X.J. Xie_2017}. It ensures that each sample point is close to one of the two hyperplanes and far away from the other hyperplane. For classification tasks, a new sample is assigned to the nearest hyperplane class. Moreover, TWSVM determines the classification hyperplane by solving two small-scale quadratic programming problems instead of one large-scale quadratic programming problem. It not only retains the advantage of SVM in handling  high-dimensional and nonlinear classification and regression problems but also achieves a training speed theoretically four times faster than SVM~\cite{X. Peng_2010}.
However, when samples are contaminated by noise, the performance of SVM and its variants will deteriorate~\cite{B.B. Gao_2015}. To reduce the uncertainty caused by outliers and noise, researchers have assigned a membership degree to each sample based on its confidence level with respect to the native class, and proposed the fuzzy support vector machine (FSVM) classifier~\cite{C.F. Lin_2002,X.W. Yang_2011,Y.T. Xu_2012,A. Shigeo_2015}. While the FSVM classifier improves the robustness, its membership function only considers the distance between the sample and the class center, and leads to confusing support vectors positioned far from the class center as noise.
In response to this problem, numerous scholars have redefined the membership function~\cite{Q. Fan_2017_Jan,S. Rezvani_2019,D. Gupta_2019_Nov,Z.Z. Liang_2022,D.A. Hua_2022,Z.Z. Liang_2024}. For instance, Fan et al. defined an entropy-based fuzzy membership function, and proposed the entropy-based fuzzy support vector machine (EFSVM) classifier~\cite{Q. Fan_2017_Jan}. Rezvani, Wang, and Pourpanah provided the intuitionistic fuzzy twin support vector machine (IFTSVM) classifier by integrating the intuitionistic fuzzy sets with the TWSVM classifier~\cite{S. Rezvani_2019}.
These studies not only reduce the impact of noise on classification tasks but also distinguish support vectors from noise.

Granular Computing (GC) is a computational paradigm that utilizes information granulation to process imprecise, inaccurate, and incomplete datasets. In 1979, Zadeh took vast amounts of information to achieve intelligent systems and controllers by leveraging the principles of GC~\cite{L.A. Zadeh_1979}. Since then, numerous scholars have integrated granular computing with machine learning~\cite{Y.Y.Yao_2009,W.P.Ding_2024,W.P.Ding_2021,Y.J. Zhang_2022,K.H. Yuan_2024,S. Ding_2015,H. Liu_2016,S. Butenkov_2017,L. Chen_2021}.
In 1982, Chen pointed out that human cognition has the characteristic of large-scale priority~\cite{L. Chen_1982}. Motivated by the cognitive mechanism of the human brain, Wang developed a framework of multi-granularity cognitive computing to handle uncertain information~\cite{G.Y. Wang_2017}. Additionally, Xia et al. provided a framework of granular-ball computing by merging granular computing and granular-balls~\cite{S.Y. Xia_2019}.
Subsequently, GBC has been widely used in various fields such as granular-ball classifiers~\cite{S.Y. Xia_2019,Y.F. Xue_2022,S.Y. Xia_2022_Oct,S.Y. Xia_2023_Apr}, granular-ball clustering~\cite{S.Y. Xia_2022_Jan,J. Xie_2023,D.D. Cheng_2024}, granular-ball rough sets~\cite{S.Y. Xia_2022_Mar,Q.H. Zhang_2023,X. Ji_2023,S.Y. Xia_2023_Nov}, and
feature selection~\cite{X.L. Peng_2022,W.B. Qian_2023_June,W.B. Qian_2023_Dec}.
Recently, Xia et al. introduced the granular-ball support vector machine (GBSVM) classifier by integrating granular-balls and the SVM classifier~\cite{S.Y. Xia_2019}.
Unlike traditional methods that use samples as inputs, the GBSVM classifier employs granular-balls as data inputs, and represents a non-point input approach with remarkable robustness and effectiveness. The robustness of GBC to noise and the efficiency of TWSVM motivate us to design a new classifier by combining granular-balls and the TWSVM classifier.
Additionally, Xue, Shao, and Xia designed the GBFSVM classifier, which exhibits higher robustness compared to the traditional FSVM classifier~\cite{Y.F. Xue_2022}.
However, the GBFSVM classifier only considers the membership degree of granular-balls to the class center, and granular-balls located in the boundary region of two classes have the same membership degree for both classes, it may lead to incorrect classifications and predictions. The current methodology solely determines granular-ball memberships by the Euclidean distance from the granular-ball's center to the class center of samples, and possibly overlooks support granular-balls situated far from class centers yet proximate to the classification boundary. It motivates us to design the GBFTSVM classifier by integrating granular-balls with the FSVM classifier. The contributions of this paper are listed as follows.

(1) We integrate GBC with the TWSVM classifier to construct the GBTWSVM classifier. This classifier uses coarse-grained granular-balls instead of traditional sample points as inputs. It ensures that the hyperplane is positioned as close as possible to one class of granular-balls while being distant from the other class of granular-balls, which enhances the robustness and efficiency of the TWSVM classifier.

(2) We combine GBC, PFS, GBRS, and FTSVM to propose the GBFTSVM classifier. This classifier introduces a new scoring function that assigns different scores to granular-balls within positive and boundary regions, and distinguishes the varied contributions of granular-balls in different regions for classification. It further enhances the performance of the GBTWSVM classifier.

(3) We perform the experiment with the GBTWSVM classifier and the GBFTSVM classifier on twenty benchmark datasets  from the UCI machine learning repository. The experimental results demonstrate that the GBTWSVM classifier and the GBFTSVM classifier outperform five mainstream classifiers in terms of accuracy, precision, recall, and running time. Particularly, the GBTWSVM classifier and the GBFTSVM classifier exhibit higher robustness to noise in the classification task.

The rest of this paper is listed as follows: Section~\ref{sec:two} reviews  GBRS, TWSVM, and FSVM. Section~\ref{sec:three} designs the GBTWSVM classifier. Section~\ref{sec:four} gives the GBFTSVM classifier. Section~\ref{sec:five} conducts the experiment with the GBTWSVM classifier and the GBFTSVM classifier. Section~\ref{sec:six} concludes this paper and shows the future work.

\section{Preliminaries}
\label{sec:two}
In this section, we review some concepts of GBC~\cite{S.Y. Xia_2019}, TWSVM~\cite{Jayadeva_2007}, and FSVM \cite{C.F. Lin_2002}.

\subsection{Granular-Ball Computing}
\label{sec:GB}
The core idea of GBC is to take a family of granular-balls to cover the original set, and use granular-balls instead of sample points as inputs. Given a dataset $U=\{(x_1,l_1),(x_2,l_2),...,(x_n,l_n)\}$, where $X=\{x_k~\vert~x_k\in \mathcal{R}^d,~k=1,2,...,n\}$ stands for the feature value matrix of $U$ with $d$ features, and $\mathcal{L}=\{l_k~\vert~l_k\in \mathcal{R},~k=1,2,...,n\}$ is the corresponding label vector. The sample universe $U$ is covered by a family of granular-balls $\mathcal{GB}=\left\{GB_i~\vert~i=1,2,...,m\right\}$, and
$c_i=\frac{1}{n_i}\sum_{k=1}^{n_i}x_{ik}$ stands for the centre of the $i$-th granular-ball $GB_i$, where $x_{ik}$ and $n_i$ stand for the $ik$-th sample and the number of samples in the $i$-th granular-ball $GB_i$, respectively. There are two methods to calculate the radius $r_i$ of the granular-ball $GB_i$: $r_i=\mathop{\max}\limits_{x_{ik}\in GB_i}\left|x_{ik}-c_i\right|$ and  $r_i=\frac{1}{n_i}\sum_{k=1}^{n_i}\left|x_{ik}-c_i\right|$.
To eliminate the effect of noisy data within each granular-ball, the overall label $y_i =\mathop{\arg \max}\limits_{l_k\in \mathcal{L}} \mid \{(x,l) \in GB_i\mid l=l_k\}\mid $ of $GB_i$ takes the label that appears most frequently within the granular-ball. The purity $p_i = \frac{| \{(x,l)\in GB_i\mid l=y_i\}|} {n_i}$ denotes the proportion of samples with the label $y_i$ in $GB_i$.

In GBC, the initial primary objective is to generate a family of granular-balls $\mathcal{GB}$, the objective function of granular-ball generation is shown as follows:
\begin{align}
\min \quad &\lambda _1\times\frac{n}{\sum\limits_{GB_{i}\in \mathcal{GB}}|GB_{i}|}+\lambda _2\times m,\\
\text{s.t.} \quad & quality(GB_i) \ge T,
\end{align}
where $\lambda _1$ and $\lambda _2$ stands for the weight coefficients, $m$ is the number of granular-balls, and $quality(GB_i)$ is the proportion of the majority of samples with the same label in the granular-ball $GB_i$.
However, the classical method of granular-ball generation faces challenges in adapting to the data distribution of each dataset. This difficulty arises from the purity threshold parameter that struggles to align with individual dataset characteristics.
To address this limitation, Xia et al. proposed a purity-adaptive method of granular-ball generation, and made the granular-ball generation completely parameter-free~\cite{S.Y. Xia_2022_Oct}. The objective function can be expressed as follows:
\begin{align}
\begin{split}
\min ~ &\lambda _1\times\frac{n}{\sum\limits_{GB_{i}\in \mathcal{GB}}|GB_{i}|}+\lambda _2\times m,\\
\text{s.t.} ~ & quality(GB_i) \ge T_0,~W(\mathcal{GB}_i)>quality(GB_i),\\
&\|c_i-c_j\|>\|r_i-r_j\|~(i,j\in[1,m],y_i\neq y_j),
\end{split}
\end{align}
where $T_0$ stands for the initial purity of the granular-ball, $\mathcal{GB}_i$ is the set of the child granular-balls of $GB_i$, and $W(\mathcal{GB}_i)$ signifies the weighted sum of purities of the child granular-balls of $GB_i$.

Suppose $S=(X,C,V,f)$ is an information system, and $\mathcal{GB}$ is a family of granular-balls towards the attribute set $Q\subseteq C$, the indiscernible granular-ball relation $INDGB(Q)$ with respect to $Q$ is defined as:
$INDGB(Q)=\{(x,z)\in X^2 \mid \exists GB \in \mathcal{GB}, x,z\in GB\}$, and $X/D=\{X_i \mid i=1,2,..,N\}$ is the partition of the universe $X$ with respect to the decision attribute set $D$, the upper and lower approximations of $X_i$ with respect to $Q$ are defined as:
$\overline{GBR_{Q}}X_i=\bigcup \{[x]_{Q} \in X/Q \mid  [x]_Q\cap X_i\ne \emptyset \};~
    \underline {GBR_{Q}}X_i=\bigcup \{[x]_{Q} \in X/Q \mid  [x]_Q\subseteq X_i \}$,
where $[x]_{Q}=\{z\in X\mid (x,z)\in INDGB(Q)\}$.
Moreover, the upper approximation and the lower approximation of $X/D$ with respect to $Q$ are defined as:
$\overline{GBR_{Q}}X =\bigcup_{i=1}^{N} \overline{GBR_{Q}}X_i;~
    \underline{GBR_{Q}}X =\bigcup_{i=1}^{N} \underline{GBR_{Q}}X_i.$
The positive region and boundary region of $X/D$ with respect to $Q$ are defined as:
\begin{align}
    POS_{Q}(X/D)&=\underline{GBR_{Q}}X;\\
    BND_{Q}(X/D)&=\overline{GBR_{Q}}X-\underline{GBR_{Q}}X.
\end{align}



\subsection{TWSVM and FSVM}
\label{sec:TWSVM}
TWSVM seeks to find a pair of non-parallel hyperplanes, and positions each hyperplane closer to homogeneous samples and somewhat far away from heterogeneous samples. A new sample $x\in\mathcal{R}^d$
is assigned to class $+1$ or $-1$
depending on which hyperplane it is closest to. A pair of non-parallel hyperplanes of the TWSVM classifier is obtained by solving the following quadratic programming problems (QPPs):
\begin{align}
\begin{split}
\underset{\omega_1,b_1,\xi_2}{\min}\quad&\frac{1}{2}(x_A\omega_1 + e_1b_1)^T(x_A\omega_1 + e_1b_1) + C_1e^T_2\xi_2,\\
\text{s.t.}\quad&-(x_B\omega_1 + e_2b_1) + \xi_2 \geq e_2, \xi_2 \geq 0,
\end{split}
\end{align}
and
\begin{align}
\begin{split}
\underset{\omega_2,b_2,\xi_1}{\min}\quad&\frac{1}{2}(x_B\omega_2 + e_2b_2)^T(x_B\omega_2 + e_2b_2) + C_2e^T_1\xi_1,\\
\text{s.t.}\quad&(x_A\omega_2 + e_1b_2) + \xi_1 \geq e_1, \xi_1 \geq 0,
\end{split}
\end{align}
where matrices $x_A$ and $x_B$ stand for the samples of classes +1 and -1, respectively, $C_1$ and $C_2$ are penalty parameters, $e_1$ and $e_2$ are vectors of ones with suitable dimensions, and $\xi_1$ and $\xi_2$ are slack variables.

A pair of non-parallel hyperplanes for the dual model of the TWSVM classifier is obtained by solving the following quadratic programming problems:
\begin{align}
\begin{split}
\underset{\alpha}{\max} \quad&\alpha^Te_2  - \frac{1}{2}\alpha^TG(H^TH)^{-1}G^T\alpha ,\\
\text{s.t.}\quad&0 \leq \alpha \leq C_1,
\end{split}
\end{align}
and
\begin{align}
\begin{split}
\underset{\gamma}{\max}\quad&\gamma^Te_1   - \frac{1}{2}\gamma^TP(Q^TQ)^{-1}P^T\gamma ,\\
\text{s.t.}\quad&0 \leq \gamma \leq C_2,
\end{split}
\end{align}
where $H = [x_A \quad e_1]$, $G = [x_B \quad e_2]$, $P = [x_A \quad e_1]$ , $Q = [x_B \quad e_2]$, $u=[\omega_1 \quad b_1]^T = (H^TH)^{-1}G^T\alpha$, and $v=[\omega_2 \quad b_2]^T = (Q^TQ)^{-1}P^T\gamma$.


Assume $\widetilde{U}=\{ (x_1,l_1,s_1),(x_2,l_2,s_2),...,(x_n,l_n,s_n)\}$ is a training dataset with fuzzy membership degrees, where $s_k\in (0,1]$ is the  membership degree that $x_k$ belongs to its label $l_k$.
The goal of FSVM is to find an optimal hyperplane by solving the following optimization problems:
\begin{align}
\begin{split}
\min \quad &\frac{1}{2}\| \omega  \|^2 +C\sum_{k = 1}^{n}s_k\xi_k,\\
\text{s.t.} \quad & l_k(\omega ^T\phi (x_k)+b)\geq 1-\xi _k,\\
&\xi _k\geq 0, k=1,2,...,n,
\end{split}
\end{align}
where $C$ is a penalty parameter, and $\xi_k$ is a slack variable.

\section{Granular-Ball Twin Support Vector Machine}
\label{sec:three}

In this section, we give the principle of the GBTWSVM classifier~\cite{L.X.Zhao_2024}.

\subsection{Granular-Ball Twin Support Vector Machine}
\label{sec:GBTWSVM}
\begin{figure}[t]
    \centering
    \includegraphics[width=0.5\textwidth]{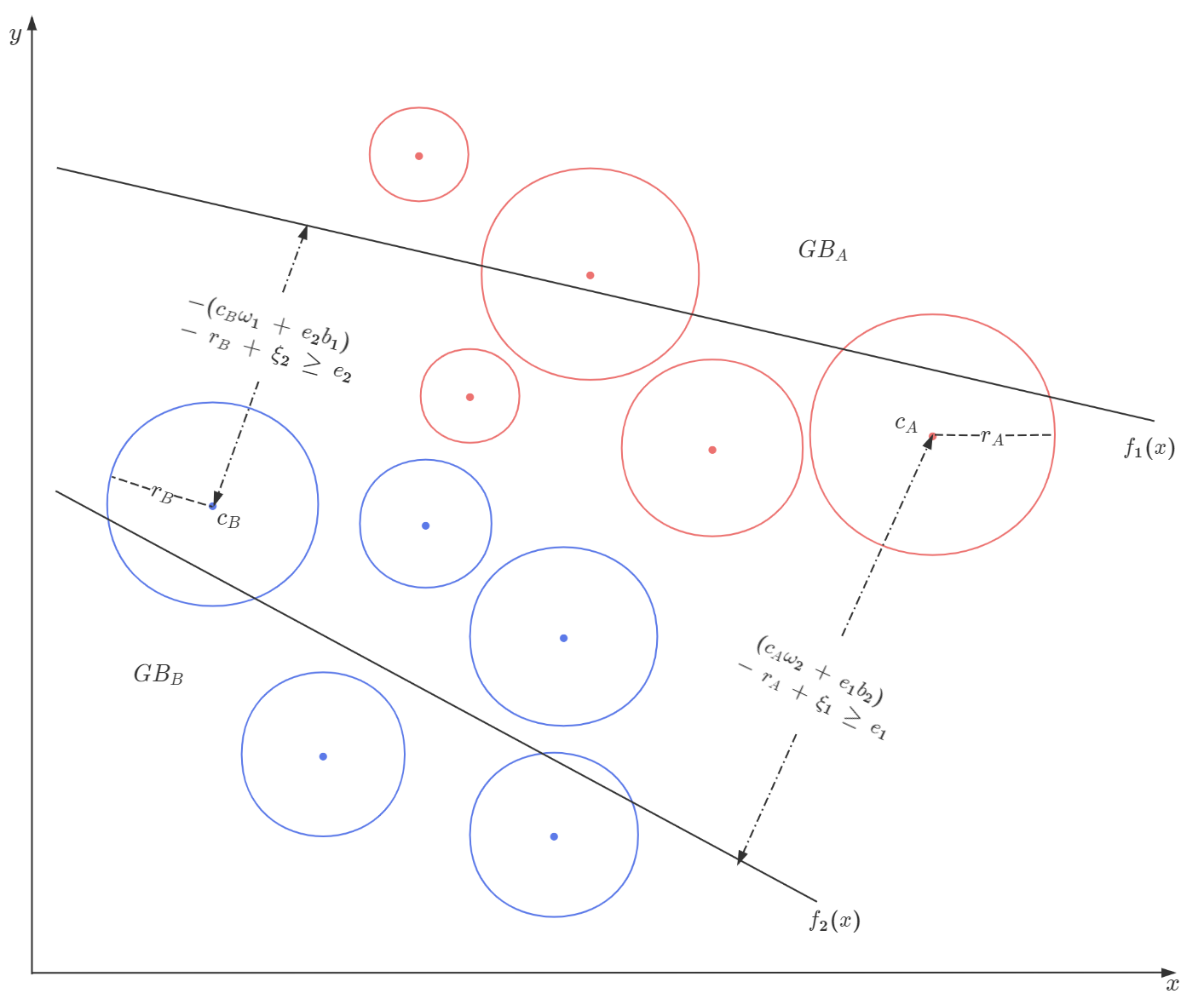}
    \caption{The GBTWSVM classifier.}
    \label{fig:GBTWSVMyuanli}
    \end{figure}
We define two non-parallel hyperplanes $f_1(x): x^T \omega _1 + b_1 = 0$ and $f_2(x): x^T \omega _2 + b_2 = 0$, where $f_1(x)$ stands for the hyperplane close to the positive-class granular-ball ($y_i=+1$), and $f_2(x)$ stands for the hyperplane close to the negative-class granular-ball ($y_i=-1$), $\omega_t$ and $b_t$ stand for the normal vector and bias of $f_t(x)$, respectively, where $t\in \{1,2\}$.
The two hyperplanes are constrained by two rules: $(1)$ the hyperplane should be as close as possible to the center of the granular-ball belonging to the same class; $(2)$ the distance between the hyperplane and the surface of granular-balls belonging to the other class should be as far as possible.
For binary classification problems, assume that $m_1$ granular-balls of class $+1$ and $m_2$ granular-balls of class $-1$ are generated, the centers of the granular-balls of classes $+1$ and $-1$ are represented by the matrices $c_A$ and $c_B$, respectively, and the radii of the granular-balls of classes $+1$ and $-1$ are represented by the matrices $r_A$ and $r_B$, respectively. For example,
we show the GBTWSVM classifier by Figure \ref{fig:GBTWSVMyuanli}, where the red and blue circles stand for granular-balls of class $+1$ and class $-1$, respectively. A pair of non-parallel hyperplanes of the GBTWSVM classifier is derived by solving the following quadratic programming problems:
\begin{align}
\begin{split}
\underset{\omega_1,b_1,\xi_2}{\min}\quad&\frac{1}{2}(c_A\omega_1 + e_1b_1)^T(c_A\omega_1 + e_1b_1) + C_1e^T_2\xi_2,\\
\text{s.t.}\quad&-(c_B\omega_1 + e_2b_1) - r_B + \xi_2 \geq e_2, \xi_2 \geq 0,
\end{split}
\end{align}
and
\begin{align}
\begin{split}
\underset{\omega_2,b_2,\xi_1}{\min}\quad&\frac{1}{2}(c_B\omega_2 + e_2b_2)^T(c_B\omega_2 + e_2b_2) + C_2e^T_1\xi_1,\\
\text{s.t.}\quad&(c_A\omega_2 + e_1b_2) - r_A + \xi_1 \geq e_1, \xi_1 \geq 0,
\end{split}
\end{align}
where $C_1$ and $C_2$ are constants and both are greater than 0, and $e_1$ and $e_2$ are unit vectors of the appropriate dimension.

\subsection{The Dual Model of the GBTWSVM classifier}
\label{sec:dual GBTWSVM}

GBTWSVM minimizes the structural risk by adding the regularization term to the margin maximization objective.
This pair of quadratic programming problems can be achieved by solving the following Lagrange function:
\begin{gather}
\begin{aligned}
\begin{split}
&L(\omega_1,b_1,\xi_2,\alpha,\beta)\\ &=\frac{1}{2}\Vert{c_A\omega_1 + e_1b_1}\Vert^2+C_1e^T_2\xi_2- \alpha^T(-(c_B\omega_1\\
&~~~+e_2b_1)+\xi_2-r_B-e_2)-\beta ^T\xi _2,\label{eq:dugbtwsvm}
\end{split}
\end{aligned}
\end{gather}
where $\alpha$ and $\beta$ are Lagrangian multipliers.

According to KKT conditions, we get:
\begin{align}
\frac{\partial L}{\partial \omega_1} = c_A^T(c_A\omega _1+e_1b_1) +c_B^T\alpha =0;\label{eq:l/w}\\
\frac{\partial L}{\partial b_1}=e_1^T(c_A\omega _1+e_1b_1) +e_2^T\alpha = 0.\label{eq:l/b}
\end{align}

By Equations~\eqref{eq:l/w} and \eqref{eq:l/b}, we obtain:
\begin{align}
\begin{bmatrix}c_A^T\\e_1^T\end{bmatrix}\begin{bmatrix}c_A&e_1\end{bmatrix}\begin{bmatrix}\omega _1\\b_1\end{bmatrix}+\begin{bmatrix}c_B^T\\e_2^T\end{bmatrix}\alpha =0.\label{eq:juzhen}
\end{align}

By taking $E = [c_A \quad e_1]$, $F = [c_B \quad e_2]$, and $u = [\omega _1 \quad b_1]^T$, Equation~\eqref{eq:juzhen} is reformulated as:
\begin{align}
E^TEu+F^T\alpha=0.
\end{align}

To improve its generalization capacity, we add the regularization item, and get the expression of $u$:
\begin{align}
u=-(E^TE+\varepsilon I)^{-1}F^T\alpha.\label{eq:u}
\end{align}

Similarly, by taking $R = [c_A \quad e_1]$, $S = [c_B \quad e_2]$, and $v = [\omega _2 \quad b_2]^T$, we get the expression of $v$:
\begin{align}
v=(S^TS+\varepsilon I)^{-1}R^T\gamma.\label{eq:v}
\end{align}

A pair of non-parallel hyperplanes for the dual model of the GBTWSVM classifier is obtained by solving the following quadratic programming problems: 
\begin{align}
\begin{split}
\underset{\alpha}{\max}\quad&\alpha^T(e_2+r_B) - \frac{1}{2}\alpha^TF(E^TE)^{-1}F^T\alpha ,\\
\text{s.t.}\quad&0 \leq \alpha \leq C_1,\label{eq:twsvm1}
\end{split}
\end{align}
and
\begin{align}
\begin{split}
\underset{\gamma}{\max}\quad&\gamma^T(e_1+r_A)  - \frac{1}{2}\gamma^TR(S^TS)^{-1}R^T\gamma,\\
\text{s.t.}\quad&0 \leq \gamma \leq C_2.\label{eq:twsvm2}
\end{split}
\end{align}

A new sample $x\,\epsilon\,\mathcal{R} ^d$ is labeled as class~$t\in\{1,2\}$ depending on which of the two hyperplanes is closer:
\begin{align}
Class~t = arg\underset{t\in \{1,2\}} {min} \frac{\left | \left \langle \omega _t,x \right \rangle +b_t \right | }{\left \| \omega _t \right \| }.
\end{align}

\subsection{Algorithm Design}

We give the algorithm for the GBTWSVM classifier as follows:

\renewcommand{\algorithmicrequire}{\textbf{Input:}}  
\renewcommand{\algorithmicensure}{\textbf{Output:}} 
\begin{algorithm}[H]
	\caption{The GBTWSVM classifier..}
	\begin{algorithmic}[1]
		\REQUIRE
		A set of granular-balls $\mathcal{GB}=\{(c_i,r_i,p_i,y_i)\mid i=1,2,...,m\}$.
		\ENSURE
		$\omega_1, \omega_2, b_1$, and $b_2$.
		\STATE Initialize $c_A, c_B, r_A$, and $r_B$ as empty matrices.
		\FOR{each $GB_i \in \mathcal{GB}$}
		\IF{$y_i=+1$}
		\STATE add $c_i$ and $r_i$ to matrices $c_A$ and $r_A$, respectively;
		\ELSE
		\STATE add $c_i$ and $r_i$ to matrices $c_B$ and $r_B$, respectively;
		\ENDIF
		\ENDFOR
		\STATE According to Equations~\eqref{eq:twsvm1} and~\eqref{eq:twsvm2}, define the objective function for optimization;
		\STATE Perform L-BFGS-B optimization for $\alpha$ and $\gamma$;
		\STATE According to Equation~\eqref{eq:u}, calculate $\omega_1$ and $b_1$;
		\STATE According to Equation~\eqref{eq:v}, calculate $\omega_2$ and $b_2$.
		\RETURN $\omega_1, \omega_2, b_1$, and $ b_2$.
	\end{algorithmic}
	\label{alg:GBTWSVM}
\end{algorithm}


Given a sample size \( n \) with approximately \( n/2 \) samples in each class, the time complexity of SVM is \( O(n^3) \). The time complexity of TWSVM is \( O(2 \times (n/2)^3) \), which is four times faster than SVM. Assume that \( n \) samples generate \( m \) granular-balls, where \( m \) is less than \( n \), and each class has approximately \( m/2 \) granular-balls, the time complexity of GBTWSVM is \( O(2 \times (m/2)^3) \). If the number of generated granular-balls \( m \) is approximately \( n/2 \), the computational speed of GBTWSVM is eight times faster than TWSVM.

\section{Granular-Ball Fuzzy Twin Support Vector Machine}
\label{sec:four}

In this section, we define the scoring function for granular-balls by utilizing Pythagorean fuzzy sets~\cite{G.M. Lang_2020}. We then assign distinct scores to granular-balls situated in the positive and boundary regions. Lastly, we present a detailed discussion on the GBFTSVM classifier.

\subsection{Pythagorean Fuzzy Membership Assignment}
\label{sec:s}

Assume $X=\left\{x_k~\vert~k=1,2,...,n\right\}$ is a dataset with $n$ sample points, we derive a set of granular-balls $\mathcal{GB}=\left\{GB_i~\vert~i=1,2,...,m\right\}$, and assign a pair of membership and non-membership degrees $(\mu_{GB_i}, \nu_{GB_i})$ to each granular-ball $GB_i$
such that $0\leq\mu_{GB_i},\nu_{GB_i}\leq 1$ and $\mu^2_{GB_i}+\nu^2_{GB_i}\leq 1$, which depicts the relationship between the granular-ball $GB_i$ and a specific class.
If $\mu_P(x_{ik})$ and $\nu_P(x_{ik})$ are the membership and non-membership degrees of a sample $x_{ik}$ in a class, respectively, the membership and non-membership degrees of the granular-ball $GB_i$ in this class are defined by:
\begin{align}
\mu_{GB_i} = \frac{1}{n_i} \sum_{k=1}^{n_i} \mu_P(x_{ik});\\
\nu_{GB_i} = \frac{1}{n_i} \sum_{k=1}^{n_i} \nu_P(x_{ik}),
\end{align}
where $x_{ik}$ and $n_i$ stand for the $ik$-th sample and the number of samples in the $i$-th granular-ball $GB_i$. In most cases, the membership and non-membership degrees of samples are unknown. So we design the membership and non-membership functions of the granular-balls as follows.

$(1)$ The membership degree: the majority of methods for constructing membership functions are based on the distance from samples to the class center. For binary-class granular-balls $\mathcal{GB}=\{(c_i, r_i, p_i, y_i)\mid 1\leq i\leq m\}$, where $c_i$, $r_i$, $p_i$, and $y_i$ represent the center, radius, purity, and label of $GB_i$, respectively, we define the class center $C^+$ and the maximum radius $R^+$ of positive-class granular-balls ($y_i=+1$), as well as the class center $C^-$ and the maximum radius $R^-$ of negative-class granular-balls ($y_i=-1$) as:
    \begin{align}
    C^+ &= \frac{1}{m_+} \sum_{y_i = +1} c_i,\\ R^+ &= \underset{y_i=+1}{\max} \left\lVert c_i-C^+ \right\rVert; \\
    C^- &= \frac{1}{m_-} \sum_{y_i = -1} c_i,\\ R^- &= \underset{y_i=-1}{\max} \left\lVert c_i-C^- \right\rVert,
    \end{align}
    where $m_+$ and $m_-$ stand for the numbers of positive and negative granular-balls, respectively. For a granular-ball $GB_i$, the membership degree can be defined as:
    \begin{align}
        \mu_{GB_i}=
        \begin{cases}
            1-\frac{\left\lVert c_i-C^+\right\rVert}{R^+ + \varepsilon}, & \text{if } y_i = +1; \\
            1-\frac{\left\lVert c_i-C^-\right\rVert}{R^- + \varepsilon}, & \text{if } y_i = -1,
        \end{cases}\label{eq:mu}
    \end{align}
    where $\varepsilon > 0$, and $\varepsilon$ is a very small number.

$(2)$ The non-membership degree: by using the granular-ball's membership degree and the granular-ball's purity, we give the non-membership function as follows:
    \begin{align}
    \nu_{GB_i}^2=(1-\mu_{GB_i}^2)(1-p_i),\label{eq:nu}
    \end{align}
    where $p_i$ stands for the purity of $GB_i$. For each granular-ball $GB_i$, the membership degree $\mu_{GB_i}$ and the non-membership degree $\nu_{GB_i}$ satisfy the following conditions: $0<\mu_{GB_i},\nu_{GB_i}\leq1$, and $0 \leq \mu_{GB_i}^2+\nu_{GB_i}^2\leq 1$.
\begin{figure}[t]
    \centering
    \includegraphics[width=0.5\textwidth]{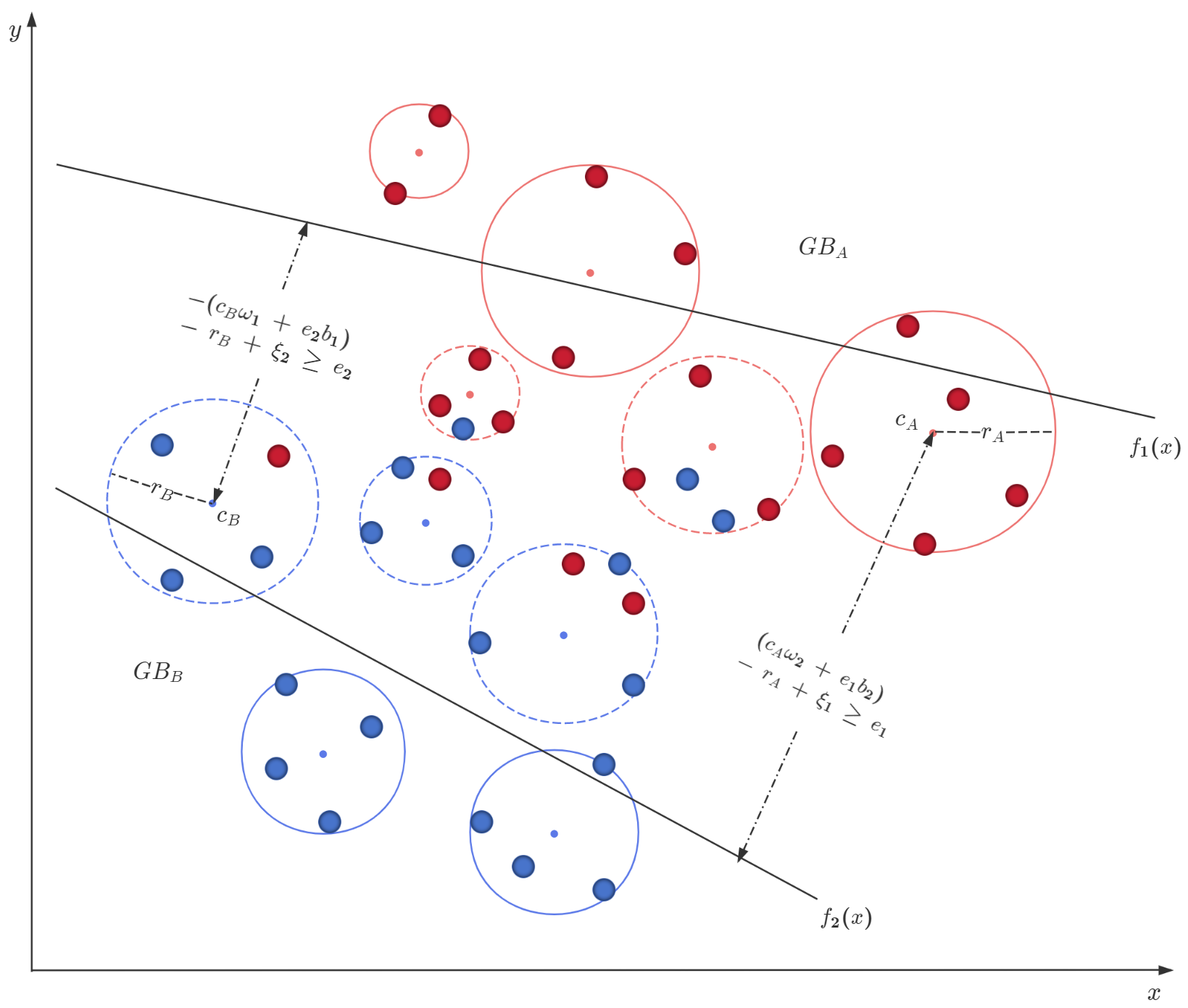}
    \caption{The GBFTSVM classifier.}
    \label{fig:GBFTSVMyuanli}
    \end{figure}

The impact of samples is not uniform. Especially, the samples in the boundary region play a crucial role in achieving accurate classifications. Conversely, the samples far from the boundary contribute relatively less to the classification. To accurately measure the contribution of granular-balls in different regions to classification, we categorize them into positive and boundary regions based on GBRS, and assign them different scoring functions. Since the overlap between heterogeneous granular-balls reduces the accuracy of classification, we employ an adaptive method of granular-ball generation to ensure that there are no heterogeneous overlaps. Based on the closeness index of Pythagorean fuzzy sets, we define the granular-ball closeness function as follows:
\begin{align}
\theta_{GB_i}=\sqrt{\frac{1-\nu_{GB_i}^2}{2-\mu_{GB_i}^2-\nu_{GB_i}^2}}.\label{eq:p}
\end{align}

Granular-balls located in the boundary region are often close to the classification hyperplane, which significantly aids in determining the decision boundary. Conversely, granular-balls in the positive region are typically far from the classification hyperplane and contribute little to determine the decision boundary. According to GBRS, if the purity of granular-ball $GB_i$ is equal to 1, then $GB_i$ belongs to the positive region. If the purity of granular-ball $GB_i$ is not equal to 1, then $GB_i$ belongs to the boundary region. Therefore, we assign a lower score $\mu_{GB_i}$ to the granular-ball in the positive region and a higher score $\theta_{GB_i}$ to that in the boundary region. Thus, the scoring function of the granular-ball $GB_i$ is defined as follows:
\begin{align}
    s_{GB_i}=
    \begin{cases}
        \mu_{GB_i}, & \text{if } p_i=1; \\
        \theta_{GB_i},  & \text{if } p_i\neq 1.
    \end{cases}
\end{align}

In Figure \ref{fig:GBFTSVMyuanli}, dashed granular-balls stand for those belonging to the boundary region, and they are typically positioned close to the separating hyperplane. We assign them a higher score. In contrast, solid granular-balls stand for those belonging to the positive region, and they are often positioned far from the separating hyperplane. This scoring function for granular-balls enables us to clearly distinguish the contribution of granular-balls in different regions.

\subsection{The Linear GBFTSVM Classifier}

Consider a set of granular-balls $\mathcal{GB}=\{(c_i, r_i, p_i, y_i)\mid 1\leq i\leq m\}$, where $c_i$ is the center of granular-ball $GB_i$, $r_i$ is the radius of granular-ball $GB_i$, $p_i$ is the purity of granular-ball $GB_i$, and $y_i\in \{-1,+1\}$ signifies the label of granular-ball $GB_i$. A pair of non-parallel hyperplanes for the linear GBFTSVM classifier is obtained by solving the following quadratic programming problems: 
\begin{align}
\begin{split}
\underset{\omega_1,b_1,\xi_2}{\min}\quad&\frac{1}{2}(c_A\omega_1 + e_1b_1)^T(c_A\omega_1 + e_1b_1) + C_1s^T_B\xi_2,\\
\text{s.t.}\quad&-(c_B\omega_1 + e_2b_1) - r_B + \xi_2 \geq e_2, \xi_2 \geq 0,
\end{split}
\end{align}
and
\begin{align}
\begin{split}
\underset{\omega_2,b_2,\xi_1}{\min}\quad&\frac{1}{2}(c_B\omega_2 + e_2b_2)^T(c_B\omega_2 + e_2b_2) + C_2s^T_A\xi_1,\\
\text{s.t.}\quad&(c_A\omega_2 + e_1b_2) - r_A + \xi_1 \geq e_1, \xi_1 \geq 0,
\end{split}
\end{align}
where $C_1$ and $C_2$ are constants and both are greater than 0, and $e_1$ and $e_2$ are unit vectors of the appropriate dimension, $s_A\in \mathcal{R} $ and $s_B\in \mathcal{R}$ are the score values of positive and negative granular-balls, respectively.

\subsection{The Dual Model of the GBFTSVM Classifier}

We tackle the quadratic programming problem of the linear GBFTSVM classifier by incorporating Lagrange multipliers $\alpha_1,~\beta_1,~\alpha_2$, and $~\beta_2$ as follows.
\begin{align}
\begin{split}
&L(\omega_1,b_1,\xi_2,\alpha_1,\beta_1)\\&=\frac{1}{2}\Vert{c_A\omega_1 + e_1b_1}\Vert^2+C_1s^T_B\xi_2- \alpha_1^T(-(c_B\omega_1+e_2b_1)\\
&~~~+\xi_2-r_B-e_2)-\beta_1 ^T\xi _2,
\end{split}
\end{align}
where $\alpha_1$ and $\beta_1$ are Lagrangian multipliers.

According to KKT conditions, we get:
\begin{align}
\frac{\partial L}{\partial \omega_1} &= c_A^T(c_A\omega _1+e_1b_1) +c_B^T\alpha =0;\label{eq:l/w1}\\
\frac{\partial L}{\partial b_1}&=e_1^T(c_A\omega _1+e_1b_1) +e_2^T\alpha = 0;\label{eq:l/b1}\\
\frac{\partial L}{\partial \xi_2} &= C_2s_B^T-\alpha _1-\beta _1.
\end{align}

By Equations~\eqref{eq:l/w1} and \eqref{eq:l/b1}, we obtain:
\begin{align}
\begin{bmatrix}c_A^T\\e_1^T\end{bmatrix}\begin{bmatrix}c_A&e_1\end{bmatrix}\begin{bmatrix}\omega _1\\b_1\end{bmatrix}+\begin{bmatrix}c_B^T\\e_2^T\end{bmatrix}\alpha_1 =0.\label{eq:juzhen1}
\end{align}

By taking $E = [c_A \quad e_1]$, $F = [c_B \quad e_2]$, and $u = [\omega _1 \quad b_1]^T$, Equation~\eqref{eq:juzhen1} is changed as:
\begin{align}
E^TEu+F^T\alpha=0.
\end{align}

Then, by adding the regularization item, we get the expression of $u$:
\begin{align}
u=-(E^TE+\varepsilon I)^{-1}F^T\alpha_1.\label{eq:u1}
\end{align}

Similarly, by taking $R = [c_A \quad e_1]$, $S = [c_B \quad e_2]$, and $v = [\omega _2 \quad b_2]^T$, we get an expression of $v$:
\begin{align}
v=(S^TS+\varepsilon I)^{-1}R^T\alpha_2.\label{eq:v1}
\end{align}

A pair of non-parallel hyperplanes for the dual model of linear GBFTSVM classifier is obtained by solving the following quadratic programming problems: 
\begin{align}
\begin{split}
\underset{\alpha_1}{\max}\quad&\alpha_1^T(e_2+r_B) - \frac{1}{2}\alpha_1^TF(E^TE)^{-1}F^T\alpha_1,\\
\text{s.t.}\quad&0 \leq \alpha_1 \leq C_3 s_B,\label{eq:gbftwsvm1}
\end{split}
\end{align}
and
\begin{align}
\begin{split}
\underset{\alpha_2}{\max}\quad&\alpha_2^T(e_1+r_A)  - \frac{1}{2}\alpha_2^TR(S^TS)^{-1}R^T\alpha_2,\\
\text{s.t.}\quad&0 \leq \alpha_2 \leq C_4 s_A.\label{eq:gbftwsvm2}
\end{split}
\end{align}

A new input data $x\,\epsilon\,\mathcal{R} ^d$ can be labeled as class~$t\in\{1,2\}$ depending on which of the two hyperplanes is closer:
\begin{align}
Class~t = arg\underset{t\in \{1,2\}} {min} \frac{\left | \left \langle \omega _t,x \right \rangle +b_t \right | }{\left \| \omega _t \right \| }.
\end{align}

\subsection{Algorithm Design}

We give the algorithm for the GBFTSVM classifier as follows:

\renewcommand{\algorithmicrequire}{\textbf{Input:}}  
\renewcommand{\algorithmicensure}{\textbf{Output:}} 
\begin{algorithm}[H]
	\caption{The GBFTSVM classifier.}
	\begin{algorithmic}[1]
		\REQUIRE
	 A set of granular-balls $\mathcal{GB}=\{(c_i,r_i,p_i,y_i)\mid i=1,2,...,m\}$.
		\ENSURE
		$\omega_1, \omega_2, b_1$, and $ b_2$.
		\STATE According to Equations~\eqref{eq:mu} and \eqref{eq:p}, calculate $\mu_{GB_i}$ and $\theta_{GB_i}$;
		\STATE Initialize $s_i$ as empty matrices.
		\FOR{each $GB_i \in \mathcal{GB}$}
		\IF {$p_i = 1$}
		\STATE add $\mu_{GB_i}$ to matrix $s_i$;
		\ELSE
		\STATE add $\theta_{GB_i}$ to matrix $s_i$;
		\ENDIF
		\ENDFOR
		\STATE Initialize $c_A,~c_B,~r_A,~r_B,~s_A$, and $s_B$ as empty matrices.
		\FOR{each $GB_i \in \mathcal{GB}$}
		\IF{$y_i=1$}
		\STATE add $c_i,~r_i$, and $s_i$ to matrices $c_A$, $r_A$, and $s_A$, respectively;
		\ELSE
		\STATE add $c_i,~r_i$, and $s_i$ to matrices $c_B$, $r_B$, and $s_B$, respectively;
		\ENDIF
		\ENDFOR
		\STATE According to Equations~\eqref{eq:gbftwsvm1} and~\eqref{eq:gbftwsvm2}, define the objective function for optimization;
		\STATE Perform L-BFGS-B optimization for $\alpha_1$ and $\alpha_2$;
		\STATE According to Equation~\eqref{eq:u1}, calculate $\omega_1$ and $b_1$;
		\STATE According to Equation~\eqref{eq:v1}, calculate $\omega_2$ and $b_2$.
		\RETURN $\omega_1, \omega_2, b_1$, and $b_2$.
		\end{algorithmic}
		\label{alg:GBFTSVM}
\end{algorithm}

%
%
%
%

Assume that \( n \) samples generate \( m \) granular-balls, where \( m \) is less than \( n \) and each class has approximately \( m/2 \) granular-balls, the time complexity of GBFTSVM is \( O\left(2\times \left(m/2\right)^3\right) \). If the number of generated granular-balls \( m \) is approximately \( n/2 \), the computational speed of GBFTSVM is eight times faster than TWSVM.

\section{Experiment Analysis}
\label{sec:five}

In this section, we compare GBFTSVM  and GBTWSVM with TWSVM~\cite{Jayadeva_2007}, EFSVM~\cite{Q. Fan_2017_Jan}, IFTSVM~\cite{S. Rezvani_2019}, GBSVM~\cite{S.Y. Xia_2019} and GBFSVM~\cite{Y.F. Xue_2022}.
All methods are implemented using Python 3.11 on a desktop with an Intel$\circledR$ Core$^{TM}$ i7-10700 CPU @ 2.90 GHz with 16 GB RAM.

\subsection{Experimental Datasets}
\label{sec:UCI Datasets}

We download twenty benchmark datasets from the UCI machine learning repository \cite{M. Kelly_2023}, and show the statistics of these benchmark datasets in Table~\ref{table:Datasets information}.
Moreover, we use ten-fold cross-validation to evaluate the performance of the seven methods in terms of $Accuracy$ ($Acc$), $Precision$ ($Prec$), $Recall$ ($Rec$), and $Standard~deviation$ of $Accuracy$ ($Sd$). To ensure a fair comparison of efficiency, the seven models are solved using the optimize library in Python under the optimal parameter settings.

\begin{table}[H]\renewcommand{\arraystretch}{1.1}
    \centering
    \caption{Datasets information.}
    \label{table:Datasets information}
    \tabcolsep0.070in
    \begin{tabular}{ccc}
    \hline
    \textbf{Datasets}            & \textbf{Numbers of samples} & \textbf{Numbers of attributes} \\ \hline
    Australian                   & 690                         & 13                             \\
    Breast-cancer                & 277                         & 9                              \\
    Breast-cancer-wisc           & 683                         & 9                              \\
    Chronic-kidney-disease       & 158                         & 24                             \\
    \begin{tabular}[c]{@{}c@{}}Congressional-voting-\\records\end{tabular} & 231                         & 16                             \\
    \begin{tabular}[c]{@{}c@{}}Conn-bench-sonar-\\ mines-rocks\end{tabular} & 208                         & 60                             \\
    Credit-approval              & 653                         & 15                             \\
    Diabetes                     & 768                         & 7                              \\
    Diabetes-upload              & 520                         & 16                             \\
    Electrical                   & 10000                       & 13                             \\
    Fourclass                    & 682                         & 2                              \\
    German.numer                 & 1000                        & 23                             \\
    Heart                        & 270                         & 12                             \\
    Ionosphere                   & 351                         & 34                             \\
    Liver-disorders              & 145                         & 4                              \\
    Messidor-features            & 1151                        & 19                             \\
    Spambase                     & 4601                        & 57                             \\
    Tic-tac-toe                  & 958                         & 8                              \\
    WDBC                         & 569                         & 30                             \\
    Wholesale-customers          & 440                         & 7                              \\ \hline
    \end{tabular}
    \end{table}

\subsection{Penalty Parameters}

\label{sec:parameters selection}
\begin{figure}
	\centering
	\begin{subfigure}[b]{0.23\textwidth}
		\centering
		\includegraphics[width=\textwidth]{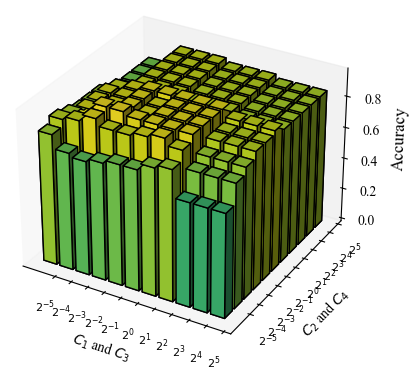}
		\captionsetup{justification=centering}
		\caption{Congressional-\\voting-records}
		\label{fig:sub1}
	\end{subfigure}
	\hfill
	\begin{subfigure}[b]{0.23\textwidth}
		\centering
		\includegraphics[width=\textwidth]{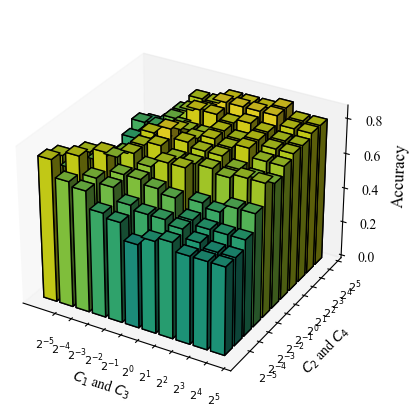}
		\captionsetup{justification=centering}
		\caption{Conn-bench-sonar-\\mines-rocks}
		\label{fig:sub2}
	\end{subfigure}
	\\
	\begin{subfigure}[b]{0.23\textwidth}
		\centering
		\includegraphics[width=\textwidth]{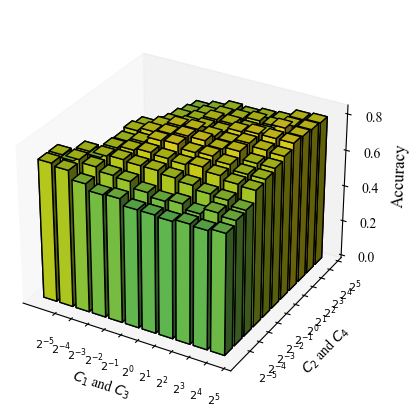}
		\caption{Diabetes}
		\label{fig:sub3}
	\end{subfigure}
	\hfill
	\begin{subfigure}[b]{0.23\textwidth}
		\centering
		\includegraphics[width=\textwidth]{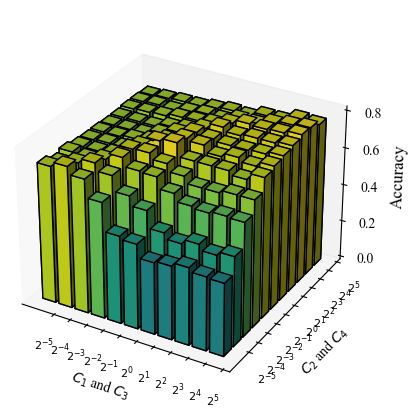}
		\caption{German.numer}
		\label{fig:sub4}
	\end{subfigure}
	\caption{$Acc$ of GBFTSVM with different $C_1$, $C_3$ and $C_2$, $C_4$.}
	\label{fig:allsub}
\end{figure}

We perform a grid search on the penalty parameters that appear in different methods, and the best results obtained are reported as the final results.
The parameters of the TWSVM classifier, the IFTSVM classifier, the GBTWSVM classifier, and the GBFTSVM classifier are set as follows: $C_i~(i=1,2,3,4)$ is explored in the grid $\{2^i \mid i=-5,-4,...,+4,+5\}$ by setting $C_1=C_3,~C_2=C_4$.
The parameters of the EFSVM classifier, the GBSVM classifier, and the GBFSVM classifier are set as follows: $C$ is explored in the grid $\{2^i \mid i=-5,-4,...,+4,+5\}$.
To investigate the impact of different parameter values on the performance of the GBFTSVM classifier, we collect the experimental results on various datasets. Due to space limitations, we select four datasets, as shown in Figure \ref{fig:allsub}. From the graph, it can be observed that the GBFTSVM classifier is highly sensitive to changes in parameter values. We select the parameters corresponding to the highest $Acc$ achieved by seven models on different datasets without noise, as shown in Table~\ref{table:parameters}.

\begin{table*}[]
    \centering
    \renewcommand{\arraystretch}{1.2}
    \caption{The optimal parameters of seven models on UCI datasets without noise.}
    \label{table:parameters}
    \begin{tabular}{lccccccc}
    \hline
    \multirow{2}{*}{\textbf{Dataset}} & TWSVM              & EFSVM    & IFTSVM             & GBSVM    & GBFSVM   & GBTWSVM            & GBFTSVM            \\
                          & $C_1$, $C_2$            & $C $       & $C_1$, $C_2$             & $C$        & $C$        &$ C_1$, $C_2 $            & $C_1$, $C_2$             \\ \hline
    Australian                                   & $2^{2}$, $2^{0}$   & $2^{-5}$ & $2^{2}$, $2^{-4}$  & $2^{-1}$ & $2^{0}$  & $2^{-4}$, $2^{-4}$ & $2^{-3}$, $2^{-5}$ \\
    Breast-cancer                                & $2^{-5}$, $2^{-4}$ & $2^{-5}$ & $2^{-5}$, $2^{-5}$ & $2^{1}$  & $2^{-1}$ & $2^{-5}$, $2^{-3}$ & $2^{4}$, $2^{-2}$  \\
    Breast-cancer-wisc                           & $2^{-5}$, $2^{-1}$ & $2^{-4}$ & $2^{-2}$, $2^{2}$  & $2^{-1}$ & $2^{0}$  & $2^{0}$, $2^{-4}$  & $2^{2}$, $2^{-2}$  \\
    Chronic-kidney-disease                       & $2^{-5}$, $2^{-5}$ & $2^{-5}$ & $2^{-5}$, $2^{-5}$ & $2^{-1}$ & $2^{-1}$ & $2^{-4}$, $2^{-5}$ & $2^{-1}$, $2^{0}$  \\
    Congressional-voting-records                 & $2^{-5}$, $2^{-5}$ & $2^{-5}$ & $2^{-5}$, $2^{-5}$ & $2^{-1}$ & $2^{-1}$ & $2^{-4}$, $2^{0}$  & $2^{-3}$, $2^{-1}$ \\
    Conn-bench-sonar-mines-rocks                 & $2^{0}$, $2^{-5}$  & $2^{5}$  & $2^{-4}$, $2^{0}$  & $2^{-2}$ & $2^{0}$  & $2^{0}$, $2^{-1}$  & $2^{3}$, $2^{2}$   \\
    Credit-approval                              & $2^{-2}$, $2^{-5}$ & $2^{-5}$ & $2^{2}$, $2^{-4}$  & $2^{-1}$ & $2^{0}$  & $2^{1}$, $2^{0}$   & $2^{0}$, $2^{3}$   \\
    Diabetes                                     & $2^{-5}$, $2^{-4}$ & $2^{-5}$ & $2^{4}$, $2^{-1}$  & $2^{-2}$ & $2^{-1}$ & $2^{3}$, $2^{-1}$  & $2^{-4}$, $2^{-5}$ \\
    Diabetes-upload                              & $2^{0}$, $2^{0}$   & $2^{-3}$ & $2^{-3}$, $2^{-3}$ & $2^{-1}$ & $2^{0}$  & $2^{-2}$, $2^{-3}$ & $2^{2}$, $2^{2}$   \\
    Electrical                                   & $2^{-4}$, $2^{-3}$ & $2^{3}$  & $2^{-1}$, $2^{-1}$ & $2^{-1}$ & $2^{0}$  & $2^{2}$, $2^{1}$   & $2^{3}$, $2^{4}$   \\
    Fourclass                                    & $2^{0}$, $2^{-1}$  & $2^{3}$  & $2^{1}$, $2^{3}$   & $2^{1}$  & $2^{-1}$ & $2^{2}$, $2^{0}$   & $2^{-5}$, $2^{-4}$ \\
    German.numer                                 & $2^{-5}$, $2^{-4}$ & $2^{-5}$ & $2^{-1}$, $2^{5}$  & $2^{1}$  & $2^{-1}$ & $2^{2}$, $2^{3}$   & $2^{-1}$, $2^{-1}$ \\
    Heart                                        & $2^{-4}$, $2^{-3}$ & $2^{-1}$ & $2^{2}$, $2^{2}$   & $2^{-1}$ & $2^{0}$  & $2^{-3}$, $2^{-5}$ & $2^{1}$, $2^{-5}$  \\
    Ionosphere                                   & $2^{-4}$, $2^{-5}$ & $2^{-5}$ & $2^{-5}$, $2^{-5}$ & $2^{-1}$ & $2^{0}$  & $2^{-5}$, $2^{3}$  & $2^{-4}$, $2^{0}$  \\
    Liver-disorders                              & $2^{-5}$, $2^{-2}$ & $2^{-1}$ & $2^{-4}$, $2^{-5}$ & $2^{0}$  & $2^{-1}$ & $2^{-1}$, $2^{-4}$ & $2^{1}$, $2^{-4}$  \\
    Messidor-features                            & $2^{-3}$, $2^{-5}$ & $2^{5}$  & $2^{0}$, $2^{-4}$  & $2^{0}$  & $2^{0}$  & $2^{3}$, $2^{4}$   & $2^{-3}$, $2^{-3}$ \\
    Spambase                                     & $2^{-5}$, $2^{-5}$ & $2^{5}$  & $2^{2}$, $2^{-2}$  & $2^{0}$  & $2^{0}$  & $2^{-2}$, $2^{-2}$ & $2^{1}$, $2^{-2}$  \\
    Tic-tac-toe                                  & $2^{-5}$, $2^{-5}$ & $2^{-5}$ & $2^{4}$, $2^{0}$   & $2^{-1}$ & $2^{-1}$ & $2^{1}$, $2^{0}$   & $2^{-2}$, $2^{-2}$ \\
    WDBC                                         & $2^{4}$, $2^{3}$   & $2^{0}$  & $2^{-4}$, $2^{-4}$ & $2^{0}$  & $2^{-1}$ & $2^{-5}$, $2^{1}$  & $2^{0}$, $2^{2}$   \\
    Wholesale-customers                          & $2^{3}$, $2^{1}$   & $2^{-5}$ & $2^{-3}$, $2^{-1}$ & $2^{0}$  & $2^{0}$  & $2^{-5}$, $2^{-4}$ & $2^{-2}$, $2^{-2}$ \\ \hline
    \end{tabular}
    \end{table*}

\begin{table*}[]
	\centering
	\caption{The running time of seven models for classification.}
	\renewcommand{\arraystretch}{1.2}
	\label{table:time}
	\begin{tabular}{lccccccc}
		\hline
		Dataset                      & TWSVM   & EFSVM    & IFTSVM   & GBSVM   & GBFSVM  & GBTWSVM        & GBFTSVM \\ \hline
		Australian                   & 2.309   & 4.178    & 5.640    & 8.758   & 6.764   & \textbf{0.172} & 0.490   \\
		Breast-cancer                & 0.289   & 0.162    & 1.495    & 1.217   & 0.638   & \textbf{0.067} & 0.100   \\
		Breast-cancer-wisc           & 2.269   & 4.541    & 5.431    & 0.149   & 0.032   & \textbf{0.016} & 0.054   \\
		Chronic-kidney-disease       & 0.450   & 0.055    & 1.061    & 0.016   & 0.046   & \textbf{0.003} & 0.009   \\
		Congressional-voting-records & 0.762   & 0.109    & 1.703    & 0.362   & 0.130   & \textbf{0.017} & 0.037   \\
		Conn-bench-sonar-mines-rocks & 0.862   & 0.102    & 0.451    & 1.067   & 0.868   & \textbf{0.032} & 0.064   \\
		Credit-approval              & 2.099   & 3.475    & 5.195    & 9.206   & 6.884   & \textbf{0.090} & 0.171   \\
		Diabetes                     & 0.762   & 5.443    & 4.202    & 5.867   & 4.382   & \textbf{0.194} & 0.274   \\
		Diabetes-upload              & 1.541   & 1.687    & 4.215    & 0.208   & 0.136   & \textbf{0.045} & 0.110   \\
		Electrical                   & 536.677 & 1511.619 & 2055.804 & 347.536 & 347.615 & \textbf{3.126} & 3.662   \\
		Fourclass                    & 1.233   & 7.462    & 7.034    & 0.156   & 0.057   & \textbf{0.034} & 0.088   \\
		German.numer                 & 1.555   & 13.508   & 6.120    & 36.314  & 22.802  & \textbf{1.038} & 1.147   \\
		Heart                        & 1.327   & 0.144    & 1.565    & 0.402   & 0.786   & \textbf{0.064} & 0.122   \\
		Ionosphere                   & 1.465   & 0.266    & 2.963    & 0.745   & 0.275   & \textbf{0.075} & 0.118   \\
		Liver-disorders              & 0.120   & 0.022    & 0.137    & 0.065   & 0.154   & \textbf{0.027} & 0.050   \\
		Messidor-features            & 2.120   & 22.039   & 6.854    & 37.686  & 42.537  & \textbf{0.940} & 1.671   \\
		Spambase                     & 99.972  & 91.115   & 651.741  & 210.924 & 201.019 & \textbf{3.129} & 3.176   \\
		Tic-tac-toe                  & 0.730   & 11.010   & 4.421    & 38.297  & 27.621  & \textbf{0.692} & 1.023   \\
		WDBC                         & 1.649   & 2.309    & 4.709    & 0.204   & 0.448   & \textbf{0.040} & 0.112   \\
		Wholesale-customers          & 1.499   & 1.046    & 3.284    & 0.366   & 0.356   & \textbf{0.107} & 0.112   \\ \hline
	\end{tabular}
\end{table*}
\subsection{Experimental Results on Datasets without Noise}

\begin{table*}[]
    \centering
    \caption{The performance of seven models on UCI datasets without noise.}
    \renewcommand{\arraystretch}{1.2}
    \label{table:acc}
    \begin{tabular}{lccccccc}
    \hline
    \multirow{3}{*}{\textbf{Dataset}} & \multicolumn{1}{c}{TWSVM}                                                       & \multicolumn{1}{c}{EFSVM}                                                       & \multicolumn{1}{c}{IFTSVM}                                                          & \multicolumn{1}{c}{GBSVM}                                                       & \multicolumn{1}{c}{GBFSVM}                                                      & \multicolumn{1}{c}{GBTWSVM}                                                     & \multicolumn{1}{c}{GBFTSVM}                                                                  \\
                            & \multicolumn{1}{c}{\begin{tabular}[c]{@{}c@{}}Acc$\pm$Sd\\ (Prec,Rec)\end{tabular}} & \multicolumn{1}{c}{\begin{tabular}[c]{@{}c@{}}Acc$\pm$Sd\\ (Prec,Rec)\end{tabular}} & \multicolumn{1}{c}{\begin{tabular}[c]{@{}c@{}}Acc$\pm$Sd\\ (Prec,Rec)\end{tabular}}     & \multicolumn{1}{c}{\begin{tabular}[c]{@{}c@{}}Acc$\pm$Sd\\ (Prec,Rec)\end{tabular}} & \multicolumn{1}{c}{\begin{tabular}[c]{@{}c@{}}Acc$\pm$Sd\\ (Prec,Rec)\end{tabular}} & \multicolumn{1}{c}{\begin{tabular}[c]{@{}c@{}}Acc$\pm$Sd\\ (Prec,Rec)\end{tabular}} & \multicolumn{1}{c}{\begin{tabular}[c]{@{}c@{}}Acc$\pm$Sd\\ (Prec,Rec)\end{tabular}}              \\ \hline
    Australian                                   & \begin{tabular}[c]{@{}l@{}}0.90$\pm$0.03\\ (0.90,0.90)\end{tabular}                 & \begin{tabular}[c]{@{}l@{}}0.84$\pm$0.08\\ (0.83,0.84)\end{tabular}                 & \begin{tabular}[c]{@{}l@{}}0.90$\pm$0.01\\ (0.90,0.90)\end{tabular}                     & \begin{tabular}[c]{@{}l@{}}0.89$\pm$0.04\\ (0.89,0.89)\end{tabular}                 & \begin{tabular}[c]{@{}l@{}}0.84$\pm$0.04\\ (0.85,0.84)\end{tabular}                 & \begin{tabular}[c]{@{}l@{}}0.91$\pm$0.03\\ (0.91,0.91)\end{tabular}                 & \textbf{\begin{tabular}[c]{@{}l@{}}0.92$\pm$0.02\\ (0.92,0.92)\end{tabular}}                     \\
    Breast-cancer                                & \begin{tabular}[c]{@{}l@{}}0.84$\pm$0.05\\ (0.83,0.84)\end{tabular}                 & \begin{tabular}[c]{@{}l@{}}0.84$\pm$0.05\\ (0.76,0.84)\end{tabular}                 & \begin{tabular}[c]{@{}l@{}}0.86$\pm$0.06\\ (0.85,0.86)\end{tabular}                     & \begin{tabular}[c]{@{}l@{}}0.86$\pm$0.07\\ (0.73,0.86)\end{tabular}                 & \begin{tabular}[c]{@{}l@{}}0.87$\pm$0.09\\ (0.88,0.87)\end{tabular}                 & \begin{tabular}[c]{@{}l@{}}0.91$\pm$0.04\\ (0.91,0.91)\end{tabular}                 & \textbf{\begin{tabular}[c]{@{}l@{}}0.92$\pm$0.03\\ (0.93,0.92)\end{tabular}}                     \\
    Breast-cancer-wisc                           & \textbf{\begin{tabular}[c]{@{}l@{}}0.99$\pm$0.01\\ (0.99,0.99)\end{tabular}}        & \begin{tabular}[c]{@{}l@{}}0.98$\pm$0.02\\ (0.98,0.98)\end{tabular}                 & \textbf{\begin{tabular}[c]{@{}l@{}}0.99$\pm$0.01\\ (0.99,0.99)\end{tabular}}            & \begin{tabular}[c]{@{}l@{}}0.98$\pm$0.02\\ (0.98,0.98)\end{tabular}                 & \begin{tabular}[c]{@{}l@{}}0.98$\pm$0.02\\ (0.98,0.98)\end{tabular}                 & \begin{tabular}[c]{@{}l@{}}0.98$\pm$0.02\\ (0.98,0.98)\end{tabular}                 & \textbf{\begin{tabular}[c]{@{}l@{}}0.99$\pm$0.01\\ (0.99,0.99)\end{tabular}}                     \\
    Chronic-kidney-disease                       & \textbf{\begin{tabular}[c]{@{}l@{}}1.00$\pm$0.00\\ (1.00,1.00)\end{tabular}}        & \begin{tabular}[c]{@{}l@{}}0.99$\pm$0.03\\ (0.99,0.99)\end{tabular}                 & \textbf{\begin{tabular}[c]{@{}l@{}}1.00$\pm$0.00\\ (1.00,1.00)\end{tabular}}            & \textbf{\begin{tabular}[c]{@{}l@{}}1.00$\pm$0.00\\ (1.00,1.00)\end{tabular}}        & \begin{tabular}[c]{@{}l@{}}0.96$\pm$0.21\\ (0.96,0.96)\end{tabular}                 & \begin{tabular}[c]{@{}l@{}}0.98$\pm$0.03\\ (0.98,0.98)\end{tabular}                 & \textbf{\begin{tabular}[c]{@{}l@{}}1.00$\pm$0.00\\ (1.00,1.00)\end{tabular}}                     \\
    Congressional-voting-records                 & \begin{tabular}[c]{@{}l@{}}0.98$\pm$0.02\\ (0.98,0.98)\end{tabular}                 & \begin{tabular}[c]{@{}l@{}}0.98$\pm$0.03\\ (0.98,0.98)\end{tabular}                 & \textbf{\begin{tabular}[c]{@{}l@{}}0.99$\pm$0.02\\ (0.99,0.99)\end{tabular}}            & \begin{tabular}[c]{@{}l@{}}0.91$\pm$0.06\\ (0.92,0.91)\end{tabular}                 & \begin{tabular}[c]{@{}l@{}}0.92$\pm$0.05\\ (0.92,0.92)\end{tabular}                 & \begin{tabular}[c]{@{}l@{}}0.97$\pm$0.04\\ (0.97,0.97)\end{tabular}                 & \textbf{\begin{tabular}[c]{@{}l@{}}0.99$\pm$0.02\\ (0.99,0.99)\end{tabular}}                     \\
    Conn-bench-sonar-mines-rocks                 & \begin{tabular}[c]{@{}l@{}}0.85$\pm$0.04\\ (0.86,0.85)\end{tabular}                 & \begin{tabular}[c]{@{}l@{}}0.80$\pm$0.08\\ (0.82,0.80)\end{tabular}                 & \begin{tabular}[c]{@{}l@{}}0.79$\pm$0.05\\ (0.80,0.79)\end{tabular}                     & \begin{tabular}[c]{@{}l@{}}0.74$\pm$0.09\\ (0.74,0.74)\end{tabular}                 & \begin{tabular}[c]{@{}l@{}}0.71$\pm$0.10\\ (0.77,0.71)\end{tabular}                 & \begin{tabular}[c]{@{}l@{}}0.93$\pm$0.03\\ (0.94,0.93)\end{tabular}                 & \textbf{\begin{tabular}[c]{@{}l@{}}0.97$\pm$0.04\\ (0.97,0.97)\end{tabular}}                     \\
    Credit-approval                              & \begin{tabular}[c]{@{}l@{}}0.91$\pm$0.03\\ (0.92,0.91)\end{tabular}                 & \begin{tabular}[c]{@{}l@{}}0.82$\pm$0.11\\ (0.79,0.82)\end{tabular}                 & \begin{tabular}[c]{@{}l@{}}0.89$\pm$0.02\\ (0.89,0.89)\end{tabular}                     & \begin{tabular}[c]{@{}l@{}}0.86$\pm$0.03\\ (0.87,0.86)\end{tabular}                 & \begin{tabular}[c]{@{}l@{}}0.88$\pm$0.05\\ (0.89,0.88)\end{tabular}                 & \begin{tabular}[c]{@{}l@{}}0.89$\pm$0.02\\ (0.90,0.89)\end{tabular}                 & \textbf{\begin{tabular}[c]{@{}l@{}}0.92$\pm$0.03\\ (0.92,0.92)\end{tabular}}                     \\
    Diabetes                                     & \begin{tabular}[c]{@{}l@{}}0.81$\pm$0.03\\ (0.82,0.81)\end{tabular}                 & \begin{tabular}[c]{@{}l@{}}0.65$\pm$0.16\\ (0.71,0.65)\end{tabular}                 & \begin{tabular}[c]{@{}l@{}}0.76$\pm$0.02\\ (0.76,0.76)\end{tabular}                     & \begin{tabular}[c]{@{}l@{}}0.79$\pm$0.03\\ (0.78,0.79)\end{tabular}                 & \begin{tabular}[c]{@{}l@{}}0.76$\pm$0.05\\ (0.78,0.76)\end{tabular}                 & \begin{tabular}[c]{@{}l@{}}0.83$\pm$0.02\\ (0.83,0.83)\end{tabular}                 & \textbf{\begin{tabular}[c]{@{}l@{}}0.84$\pm$0.03\\ (0.85,0.84)\end{tabular}}                     \\
    Diabetes-upload                              & \begin{tabular}[c]{@{}l@{}}0.96$\pm$0.03\\ (0.96,0.96)\end{tabular}                 & \begin{tabular}[c]{@{}l@{}}0.92$\pm$0.03\\ (0.93,0.92)\end{tabular}                 & \begin{tabular}[c]{@{}l@{}}0.94$\pm$0.04\\ (0.94,0.94)\end{tabular}                     & \begin{tabular}[c]{@{}l@{}}0.84$\pm$0.15\\ (0.86,0.84)\end{tabular}                 & \begin{tabular}[c]{@{}l@{}}0.82$\pm$0.08\\ (0.83,0.82)\end{tabular}                 & \begin{tabular}[c]{@{}l@{}}0.96$\pm$0.02\\ (0.96,0.96)\end{tabular}                 & \textbf{\begin{tabular}[c]{@{}l@{}}0.98$\pm$0.02\\ (0.98,0.98)\end{tabular}}                     \\
    Electrical                                   & \begin{tabular}[c]{@{}l@{}}0.99$\pm$0.00\\ (0.99,0.99)\end{tabular}                 & \textbf{\begin{tabular}[c]{@{}l@{}}1.00$\pm$0.00\\ (1.00,1.00)\end{tabular}}        & \multicolumn{1}{c}{\begin{tabular}[c]{@{}c@{}}0.98$\pm$0.01\\ (0.98,0.98)\end{tabular}} & \begin{tabular}[c]{@{}l@{}}0.75$\pm$0.03\\ (0.76,0.75)\end{tabular}                 & \begin{tabular}[c]{@{}l@{}}0.80$\pm$0.07\\ (0.82,0.80)\end{tabular}                 & \textbf{\begin{tabular}[c]{@{}l@{}}1.00$\pm$0.00\\ (1.00,1.00)\end{tabular}}        & \textbf{\begin{tabular}[c]{@{}l@{}}1.00$\pm$0.00\\ (1.00,1.00)\end{tabular}}                     \\
    Fourclass                                    & \begin{tabular}[c]{@{}l@{}}0.79$\pm$0.04\\ (0.81,0.79)\end{tabular}                 & \begin{tabular}[c]{@{}l@{}}0.74$\pm$0.06\\ (0.81,0.74)\end{tabular}                 & \begin{tabular}[c]{@{}l@{}}0.82$\pm$0.03\\ (0.83,0.82)\end{tabular}                     & \begin{tabular}[c]{@{}l@{}}0.83$\pm$0.04\\ (0.84,0.83)\end{tabular}                 & \begin{tabular}[c]{@{}l@{}}0.76$\pm$0.04\\ (0.78,0.76)\end{tabular}                 & \begin{tabular}[c]{@{}l@{}}0.83$\pm$0.03\\ (0.85,0.83)\end{tabular}                 & \textbf{\begin{tabular}[c]{@{}l@{}}0.84$\pm$0.03\\ (0.85,0.84)\end{tabular}}                     \\
    German.numer                                 & \begin{tabular}[c]{@{}l@{}}0.77$\pm$0.03\\ (0.77,0.77)\end{tabular}                 & \begin{tabular}[c]{@{}l@{}}0.75$\pm$0.05\\ (0.71,0.75)\end{tabular}                 & \begin{tabular}[c]{@{}l@{}}0.76$\pm$0.03\\ (0.76,0.76)\end{tabular}                     & \begin{tabular}[c]{@{}l@{}}0.80$\pm$0.04\\ (0.79,0.80)\end{tabular}                 & \begin{tabular}[c]{@{}l@{}}0.80$\pm$0.04\\ (0.82,0.80)\end{tabular}                 & \begin{tabular}[c]{@{}l@{}}0.81$\pm$0.04\\ (0.80,0.81)\end{tabular}                 & \multicolumn{1}{c}{\textbf{\begin{tabular}[c]{@{}c@{}}0.83$\pm$0.03\\ (0.82,0.83)\end{tabular}}} \\
    Heart                                        & \begin{tabular}[c]{@{}l@{}}0.88$\pm$0.06\\ (0.89,0.88)\end{tabular}                 & \begin{tabular}[c]{@{}l@{}}0.89$\pm$0.06\\ (0.90,0.89)\end{tabular}                 & \begin{tabular}[c]{@{}l@{}}0.90$\pm$0.04\\ (0.90,0.90)\end{tabular}                     & \begin{tabular}[c]{@{}l@{}}0.85$\pm$0.04\\ (0.86,0.85)\end{tabular}                 & \begin{tabular}[c]{@{}l@{}}0.83$\pm$0.08\\ (0.85,0.83)\end{tabular}                 & \begin{tabular}[c]{@{}l@{}}0.91$\pm$0.05\\ (0.91,0.91)\end{tabular}                 & \textbf{\begin{tabular}[c]{@{}l@{}}0.95$\pm$0.04\\ (0.95,0.95)\end{tabular}}                     \\
    Ionosphere                                   & \begin{tabular}[c]{@{}l@{}}0.93$\pm$0.04\\ (0.93,0.93)\end{tabular}                 & \begin{tabular}[c]{@{}l@{}}0.86$\pm$0.14\\ (0.84,0.86)\end{tabular}                 & \begin{tabular}[c]{@{}l@{}}0.95$\pm$0.03\\ (0.95,0.95)\end{tabular}                     & \begin{tabular}[c]{@{}l@{}}0.80$\pm$0.04\\ (0.83,0.80)\end{tabular}                 & \begin{tabular}[c]{@{}l@{}}0.78$\pm$0.06\\ (0.74,0.78)\end{tabular}                 & \begin{tabular}[c]{@{}l@{}}0.93$\pm$0.02\\ (0.94,0.93)\end{tabular}                 & \textbf{\begin{tabular}[c]{@{}l@{}}0.97$\pm$0.02\\ (0.97,0.97)\end{tabular}}                     \\
    Liver-disorders                              & \begin{tabular}[c]{@{}l@{}}0.87$\pm$0.12\\ (0.86,0.87)\end{tabular}                 & \begin{tabular}[c]{@{}l@{}}0.75$\pm$0.07\\ (0.82,0.75)\end{tabular}                 & \begin{tabular}[c]{@{}l@{}}0.85$\pm$0.08\\ (0.87,0.85)\end{tabular}                     & \begin{tabular}[c]{@{}l@{}}0.86$\pm$0.12\\ (0.88,0.86)\end{tabular}                 & \begin{tabular}[c]{@{}l@{}}0.79$\pm$0.10\\ (0.78,0.79)\end{tabular}                 & \begin{tabular}[c]{@{}l@{}}0.90$\pm$0.05\\ (0.91,0.90)\end{tabular}                 & \textbf{\begin{tabular}[c]{@{}l@{}}0.93$\pm$0.04\\ (0.94,0.93)\end{tabular}}                     \\
    Messidor-features                            & \begin{tabular}[c]{@{}l@{}}0.80$\pm$0.03\\ (0.82,0.80)\end{tabular}                 & \begin{tabular}[c]{@{}l@{}}0.70$\pm$0.05\\ (0.76,0.70)\end{tabular}                 & \begin{tabular}[c]{@{}l@{}}0.76$\pm$0.02\\ (0.78,0.76)\end{tabular}                     & \begin{tabular}[c]{@{}l@{}}0.70$\pm$0.05\\ (0.70,0.70)\end{tabular}                 & \begin{tabular}[c]{@{}l@{}}0.71$\pm$0.04\\ (0.71,0.71)\end{tabular}                 & \textbf{\begin{tabular}[c]{@{}l@{}}0.82$\pm$0.03\\ (0.82,0.82)\end{tabular}}        & \multicolumn{1}{c}{\begin{tabular}[c]{@{}c@{}}0.80$\pm$0.03\\ (0.82,0.80)\end{tabular}}          \\
    Spambase                                     & \begin{tabular}[c]{@{}l@{}}0.85$\pm$0.02\\ (0.87,0.85)\end{tabular}                 & \begin{tabular}[c]{@{}l@{}}0.91$\pm$0.04\\ (0.91,0.91)\end{tabular}                 & \begin{tabular}[c]{@{}l@{}}0.88$\pm$0.02\\ (0.89,0.88)\end{tabular}                     & \begin{tabular}[c]{@{}l@{}}0.66$\pm$0.02\\ (0.67,0.66)\end{tabular}                 & \begin{tabular}[c]{@{}l@{}}0.60$\pm$0.13\\ (0.60,0.60)\end{tabular}                 & \textbf{\begin{tabular}[c]{@{}l@{}}0.94$\pm$0.01\\ (0.94,0.94)\end{tabular}}        & \begin{tabular}[c]{@{}l@{}}0.93$\pm$0.00\\ (0.93,0.93)\end{tabular}                              \\
    Tic-tac-toe                                  & \begin{tabular}[c]{@{}l@{}}0.70$\pm$0.04\\ (0.68,0.70)\end{tabular}                 & \begin{tabular}[c]{@{}l@{}}0.56$\pm$0.13\\ (0.33,0.56)\end{tabular}                 & \begin{tabular}[c]{@{}l@{}}0.70$\pm$0.04\\ (0.73,0.70)\end{tabular}                     & \begin{tabular}[c]{@{}l@{}}0.67$\pm$0.06\\ (0.68,0.67)\end{tabular}                 & \begin{tabular}[c]{@{}l@{}}0.68$\pm$0.04\\ (0.55,0.68)\end{tabular}                 & \begin{tabular}[c]{@{}l@{}}0.75$\pm$0.05\\ (0.76,0.75)\end{tabular}                 & \textbf{\begin{tabular}[c]{@{}l@{}}0.77$\pm$0.03\\ (0.80,0.77)\end{tabular}}                     \\
    WDBC                                         & \begin{tabular}[c]{@{}l@{}}0.98$\pm$0.01\\ (0.98,0.98)\end{tabular}                 & \begin{tabular}[c]{@{}l@{}}0.97$\pm$0.02\\ (0.98,0.97)\end{tabular}                 & \textbf{\begin{tabular}[c]{@{}l@{}}1.00$\pm$0.01\\ (1.00,1.00)\end{tabular}}            & \begin{tabular}[c]{@{}l@{}}0.96$\pm$0.03\\ (0.96,0.96)\end{tabular}                 & \begin{tabular}[c]{@{}l@{}}0.87$\pm$0.15\\ (0.85,0.87)\end{tabular}                 & \begin{tabular}[c]{@{}l@{}}0.96$\pm$0.03\\ (0.97,0.96)\end{tabular}                 & \begin{tabular}[c]{@{}l@{}}0.99$\pm$0.01\\ (0.99,0.99)\end{tabular}                              \\
    Wholesale-customers                          & \begin{tabular}[c]{@{}l@{}}0.95$\pm$0.04\\ (0.95,0.95)\end{tabular}                 & \begin{tabular}[c]{@{}l@{}}0.85$\pm$0.21\\ (0.84,0.85)\end{tabular}                 & \begin{tabular}[c]{@{}l@{}}0.94$\pm$0.03\\ (0.94,0.94)\end{tabular}                     & \begin{tabular}[c]{@{}l@{}}0.85$\pm$0.05\\ (0.85,0.85)\end{tabular}                 & \begin{tabular}[c]{@{}l@{}}0.84$\pm$0.04\\ (0.85,0.84)\end{tabular}                 & \begin{tabular}[c]{@{}l@{}}0.95$\pm$0.04\\ (0.95,0.95)\end{tabular}                 & \textbf{\begin{tabular}[c]{@{}l@{}}0.96$\pm$0.02\\ (0.97,0.96)\end{tabular}}                     \\ \hline
    Average Performance                                  & \begin{tabular}[c]{@{}l@{}}0.89$\pm$0.03\\ (0.89,0.89)\end{tabular}                & \begin{tabular}[c]{@{}l@{}}0.84$\pm$0.07\\ (0.83,0.84)\end{tabular}          & \begin{tabular}[c]{@{}l@{}}0.88$\pm$0.03\\ (0.89,0.88)\end{tabular}                             & \begin{tabular}[c]{@{}l@{}}0.83$\pm$0.05\\ (0.83,0.83)\end{tabular}                & \begin{tabular}[c]{@{}l@{}}0.81$\pm$0.07\\ (0.81,0.81)\end{tabular}                  & \begin{tabular}[c]{@{}l@{}}0.91$\pm$0.03\\ (0.91,0.91)\end{tabular}                & \textbf{\begin{tabular}[c]{@{}l@{}}0.93$\pm$0.02\\ (0.93,0.93)\end{tabular}}                                                                   \\ \hline
    Average Rank of Acc                                 & \multicolumn{1}{c}{3.4}                                                         & \multicolumn{1}{c}{5.175}                                                       & \multicolumn{1}{c}{3.55}                                                            & \multicolumn{1}{c}{5.05}                                                        & \multicolumn{1}{c}{6.525}                                                       & \multicolumn{1}{c}{2.825}                                                       & \multicolumn{1}{c}{\textbf{1.35}}                                                                    \\ \hline
    \end{tabular}
    \end{table*}

\begin{table*}[http]
	\centering
	\caption{The performance of seven models on UCI datasets with different noise levels.}
	\renewcommand{\arraystretch}{1.1}
	\label{table:noise}
	\begin{tabular}{lcccccccc}
		\hline
		\multirow{2}{*}{Dataset}     & \multirow{2}{*}{Noisy} & TWSVM         & EFSVM         & IFTSVM        & GBSVM         & GBFSVM        & GBTWSVM       & GBFTSVM       \\
		&                        & Acc           & Acc           & Acc           & Acc           & Acc           & Acc           & Acc           \\ \hline
		Australian                   & 0                      & 0.90          & 0.84          & 0.90          & 0.89          & 0.84          & 0.91          & \textbf{0.92} \\
		& 0.05                   & 0.88          & 0.82          & 0.89          & 0.86          & 0.84          & 0.90          & \textbf{0.90} \\
		& 0.1                    & 0.88          & 0.88          & 0.89          & 0.89          & 0.86          & 0.89          & \textbf{0.91} \\
		Breast-cancer                & 0                      & 0.84          & 0.84          & 0.86          & 0.86          & 0.87          & 0.91          & \textbf{0.92} \\
		& 0.05                   & 0.80          & 0.79          & 0.84          & 0.78          & 0.82          & \textbf{0.90} & \textbf{0.90} \\
		& 0.1                    & 0.81          & 0.83          & 0.86          & 0.80          & 0.88          & 0.88          & \textbf{0.90} \\
		Breast-cancer-wisc           & 0                      & 0.99          & 0.98          & \textbf{0.99} & 0.98          & 0.98          & 0.98          & \textbf{0.99} \\
		& 0.05                   & 0.98          & 0.95          & 0.98          & 0.98          & 0.98          & 0.99          & \textbf{0.99} \\
		& 0.1                    & 0.98          & 0.93          & 0.97          & 0.98          & 0.98          & 0.99          & \textbf{0.99} \\
		Chronic-kidney-disease       & 0                      & \textbf{1.00} & 0.99          & \textbf{1.00} & \textbf{1.00} & 0.96          & 0.98          & \textbf{1.00} \\
		& 0.05                   & 0.99          & \textbf{1.00} & \textbf{1.00} & \textbf{1.00} & \textbf{1.00} & 0.99          & \textbf{1.00} \\
		& 0.1                    & 0.99          & \textbf{1.00} & \textbf{1.00} & \textbf{1.00} & 0.99          & \textbf{1.00} & 0.99          \\
		Congressional-voting-records & 0                      & 0.98          & 0.98          & \textbf{0.99} & 0.91          & 0.92          & 0.97          & \textbf{0.99} \\
		& 0.05                   & \textbf{0.98} & 0.97          & \textbf{0.98} & 0.90          & 0.90          & 0.98          & \textbf{0.98} \\
		& 0.1                    & 0.97          & 0.95          & 0.98          & 0.91          & 0.93          & 0.98          & \textbf{0.99} \\
		Conn-bench-sonar-mines-rocks & 0                      & 0.85          & 0.80          & 0.79          & 0.74          & 0.71          & 0.93          & \textbf{0.97} \\
		& 0.05                   & 0.83          & 0.83          & 0.79          & 0.78          & 0.60          & 0.90          & \textbf{0.92} \\
		& 0.1                    & 0.80          & 0.75          & 0.76          & 0.72          & 0.61          & 0.90          & \textbf{0.90} \\
		Credit-approval              & 0                      & 0.91          & 0.82          & 0.89          & 0.86          & 0.88          & 0.89          & \textbf{0.92} \\
		& 0.05                   & 0.90          & 0.77          & 0.88          & 0.88          & 0.88          & 0.89          & \textbf{0.90} \\
		& 0.1                    & 0.89          & 0.87          & 0.89          & 0.85          & 0.87          & 0.90          & \textbf{0.92} \\
		Diabetes                     & 0                      & 0.81          & 0.65          & 0.76          & 0.79          & 0.76          & 0.83          & \textbf{0.84} \\
		& 0.05                   & 0.81          & 0.62          & 0.80          & 0.77          & 0.75          & 0.84          & \textbf{0.85} \\
		& 0.1                    & 0.79          & 0.47          & 0.76          & 0.76          & 0.74          & 0.83          & \textbf{0.84} \\
		Diabetes-upload              & 0                      & 0.96          & 0.92          & 0.94          & 0.84          & 0.82          & 0.96          & \textbf{0.98} \\
		& 0.05                   & 0.94          & 0.83          & 0.94          & 0.83          & 0.90          & \textbf{0.96} & \textbf{0.96} \\
		& 0.1                    & 0.93          & 0.86          & 0.93          & 0.83          & 0.82          & 0.97          & \textbf{0.97} \\
		Electrical                   & 0                      & 0.99          & \textbf{1.00} & 0.98          & 0.75          & 0.80          & \textbf{1.00} & \textbf{1.00} \\
		& 0.05                   & 0.93          & 0.97          & 0.96          & 0.74          & 0.81          & \textbf{1.00} & \textbf{1.00} \\
		& 0.1                    & 0.89          & 0.90          & 0.93          & 0.73          & 0.80          & \textbf{0.99} & \textbf{0.99} \\
		Fourclass                    & 0                      & 0.79          & 0.74          & 0.82          & 0.83          & 0.76          & 0.83          & \textbf{0.84} \\
		& 0.05                   & 0.80          & 0.77          & 0.78          & 0.81          & 0.73          & 0.82          & \textbf{0.83} \\
		& 0.1                    & 0.79          & 0.71          & 0.79          & 0.81          & 0.73          & \textbf{0.82} & \textbf{0.82} \\
		German.numer                 & 0                      & 0.77          & 0.75          & 0.76          & 0.80          & 0.80          & 0.81          & \textbf{0.83} \\
		& 0.05                   & 0.78          & 0.79          & 0.78          & 0.81          & 0.78          & 0.82          & \textbf{0.83} \\
		& 0.1                    & 0.77          & 0.76          & 0.76          & 0.81          & 0.78          & 0.80          & \textbf{0.82} \\
		Heart                        & 0                      & 0.88          & 0.89          & 0.90          & 0.85          & 0.83          & 0.91          & \textbf{0.95} \\
		& 0.05                   & 0.88          & 0.81          & 0.86          & 0.82          & 0.86          & 0.91          & \textbf{0.92} \\
		& 0.1                    & 0.86          & 0.81          & 0.87          & 0.83          & 0.81          & 0.91          & \textbf{0.93} \\
		Ionosphere                   & 0                      & 0.93          & 0.86          & 0.95          & 0.80          & 0.78          & 0.93          & \textbf{0.97} \\
		& 0.05                   & 0.93          & 0.76          & \textbf{0.94} & 0.78          & 0.81          & 0.93          & \textbf{0.94} \\
		& 0.1                    & 0.91          & 0.71          & 0.94          & 0.78          & 0.78          & 0.93          & \textbf{0.96} \\
		Liver-disorders              & 0                      & 0.87          & 0.75          & 0.85          & 0.86          & 0.79          & 0.90          & \textbf{0.93} \\
		& 0.05                   & 0.83          & 0.83          & 0.83          & 0.83          & 0.73          & 0.90          & \textbf{0.94} \\
		& 0.1                    & 0.81          & 0.75          & 0.81          & 0.82          & 0.59          & 0.88          & \textbf{0.93} \\
		Messidor-features            & 0                      & 0.80          & 0.70          & 0.76          & 0.70          & 0.71          & \textbf{0.82} & 0.80          \\
		& 0.05                   & 0.78          & 0.69          & 0.75          & 0.70          & 0.70          & \textbf{0.80} & 0.79          \\
		& 0.1                    & 0.76          & 0.67          & 0.73          & 0.69          & 0.71          & 0.77          & \textbf{0.78} \\
		Spambase                     & 0                      & 0.85          & 0.91          & 0.88          & 0.66          & 0.60          & \textbf{0.94} & 0.93          \\
		& 0.05                   & 0.83          & 0.88          & 0.86          & 0.68          & 0.62          & 0.93          & \textbf{0.94} \\
		& 0.1                    & 0.80          & 0.86          & 0.84          & 0.66          & 0.61          & 0.92          & \textbf{0.94} \\
		Tic-tac-toe                  & 0                      & 0.70          & 0.56          & 0.70          & 0.67          & 0.68          & 0.75          & \textbf{0.77} \\
		& 0.05                   & 0.68          & 0.46          & 0.68          & 0.65          & 0.67          & 0.74          & \textbf{0.75} \\
		& 0.1                    & 0.67          & 0.61          & 0.63          & 0.63          & 0.63          & 0.73          & \textbf{0.73} \\
		WDBC                         & 0                      & 0.98          & 0.97          & \textbf{1.00} & 0.96          & 0.77          & 0.96          & 0.99          \\
		& 0.05                   & 0.97          & 0.97          & \textbf{0.99} & 0.94          & 0.81          & 0.97          & 0.98          \\
		& 0.1                    & 0.97          & 0.90          & \textbf{0.98} & 0.93          & 0.89          & 0.97          & \textbf{0.98} \\
		Wholesale-customers          & 0                      & 0.95          & 0.85          & 0.94          & 0.85          & 0.87          & 0.95          & \textbf{0.96} \\
		& 0.05                   & 0.93          & 0.82          & 0.94          & 0.87          & 0.82          & 0.95          & \textbf{0.96} \\
		& 0.1                    & 0.92          & 0.64          & 0.90          & 0.81          & 0.82          & \textbf{0.95} & \textbf{0.95} \\ \hline
		Average Acc                  & 0                      & 0.89          & 0.84          & 0.88          & 0.83          & 0.81          & 0.91          & \textbf{0.93} \\
		& 0.05                   & 0.87          & 0.82          & 0.87          & 0.82          & 0.80          & 0.91          & \textbf{0.91} \\
		& 0.1                    & 0.86          & 0.79          & 0.86          & 0.81          & 0.79          & 0.90          & \textbf{0.91} \\ \hline
	\end{tabular}
\end{table*}

\subsubsection{Comparison of Training Times}

In the absence of noise, we show the training time of these models across various datasets in Table~\ref{table:time}. Notably, the GBTWSVM classifier exhibits the shortest training time across all datasets, and achieves remarkable speedups of hundreds of times in certain cases. In addition to the GBTWSM classifier, the GBFTSVM classifier has the shortest training time among the remaining models.
This observation can be attributed to two primary reasons. First, the number of granular-balls generated in the datasets is significantly smaller than the sample size. Second, both GBTWSVM and GBFTSVM derive the hyperplane by two smaller-scale quadratic programming problems instead of a single large-scale one.
Furthermore, although the GBTWSVM classifier performs slightly worse $Acc$ than the GBFTSVM classifier, it exhibits faster. This is because the GBFTSVM classifier requires the computation of a scoring function for each granular-ball to differentiate their contributions to classification, whereas the GBTWSVM classifier does not incorporate this aspect.

\subsubsection{Comparison of Performance}
To comprehensively evaluate the performance of the GBTWSVM classifier and the GBFTSVM classifier on multiple datasets without noise, we conduct a rigorous comparison of seven models across 20 benchmark datasets. We evaluate these models based on their $Acc$, $Pre$, $Rec$, and $Sd$. The results are summarized in Table~\ref{table:acc}. After a thorough analysis, we arrive at the following conclusions:
the GBFTSVM classifier exhibits the highest $Acc$, $Pre$, and $Rec$ in 17 out of the 20 datasets, and firmly establishes its superior classification capabilities. Notably, the GBFTSVM classifier also gives a relatively low standard deviation in $Acc$, which indicates its consistent and reliable performance. While the GBTWSVM classifier trails the GBFTSVM classifier slightly in terms of overall $Acc$, it achieves high $Acc$ in 14 datasets with a low standard deviation, and emerges as a strong contender among the remaining six models.
Furthermore, we compare the average $Acc$ of each model across all datasets. Evidently, the GBFTSVM classifier achieves the highest average $Acc$, and outperforms the other models by 2\%-12\%, which underscores its superior classification performance. In comparison with the GBTWSVM classifier, the notable improvement of the GBFTSVM classifier's average $Acc$ validates the significance of our optimizations and enhancements to the membership function for classification.

To overcome the potential bias caused by a model's high $Acc$ in one dataset and low $Acc$s in others, we calculate the average rank of each model across the datasets. The model with the highest $Acc$ is ranked first, and in cases of equal $Acc$, the average of the corresponding ranks is taken. The GBFTSVM classifier emerges with the lowest average rank, and firmly establishes its superiority among all the compared models. Similarly, the GBTWSVM classifier's second-lowest average rank demonstrates its strong performance among the remaining models, excluding the GBFTSVM classifier.

\subsection{Experimental Results on Datasets with Noise}

To assess the robustness of the GBTWSVM classifier  and the GBFTSVM classifier  to noise, we add 5\% and 10\% label noise to $20$ datasets and compare the $Acc$ of seven models. In Table~\ref{table:noise}, the GBFTSVM classifier  gives the highest $Acc$ on 18 datasets with 5\% noise and on 19 datasets with 10\% noise, and outperforms the other six models. Notably, in the Messidor-features and WDBC datasets, although the GBFTSVM classifier  does not exhibit optimal performance in the absence of noise, its stability in the face of noise leads to improved classification performance. Especially, it surpasses other models on datasets with 10\% noise.
It is noteworthy that classifiers based on granular-balls exhibits a relatively stable $Acc$ fluctuation of approximately 2\% with noise addition, compared to a fluctuation of over 2\% for classifiers based on point inputs. This suggests that classifiers based on granular-balls offer better robustness to noise in terms of prediction stability.
Furthermore, the GBFTSVM classifier emerges as the top performer in terms of average $Acc$ across all noise levels (0\%, 5\%, and 10\%). Several factors contribute to this performance. First, the coarser size of granular-balls and the assignment of majority point labels reduce the impact of noisy points. Second, the GBFTSVM classifier  employs two symmetric support vectors based on granular-balls, which, through their symmetry, effectively resists noise and outliers, and enhances generalization ability. Third, the assignment of scoring functions to granular-balls effectively distinguishes supportive granular-balls, and minimizes their influence on determining the separation hyperplane.


\subsection{Statistical Analysis}

\begin{figure}[H]
    \centering
    \includegraphics[width=0.5\textwidth]{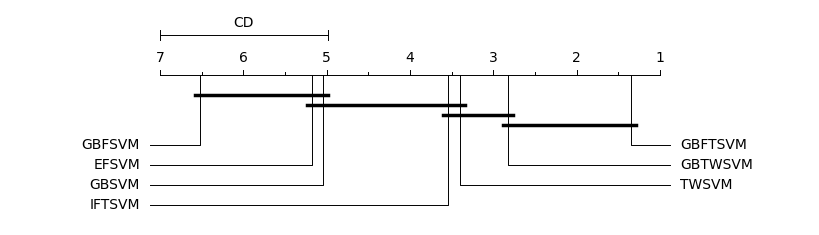}
    \caption{The comparison of various models in terms of CD diagrams.}
    \label{fig:CD}
    \end{figure}

To validate the statistical significance of the GBFTSVM classifier, we conduct Friedman test and Nemenyi post-hoc test on seven models across 20 datasets without noise. Initially, we assume that all models are equivalent under the null hypothesis. Friedman statistic follows a chi-square distribution, which is calculated using the formula:
$\chi _F ^{2} =\frac{12N}{M(M+1)}\left [ \sum_{j=1}^{M}R_j^{2}-\frac{M(M+1)^{2}}{4}   \right ], $
where $R_j$ $(j=1,2,...7)$ stands for the average rank of the $j^{th}$ model, $N$ is the number of datasets, and $M$ is the number of models. Alternatively, when Friedman statistic follows an F-distribution with $((M-1),(M-1)(M-N))$ degrees of freedom, the formula is:
$F_F=\frac{(N-1)\chi_F^{2}}{N(M-1)-\chi_F^{2}}$. Based on Table~\ref{table:acc}, we obtain an $F_F$ value of 28.6 with degrees of freedom (6,114). At a significance level of 0.05, the critical value of $F(6,114)$ is 2.179. Since $F_F$ exceeds this critical value, we reject the null hypothesis.
Subsequently, we ultilize Nemenyi post-hoc test to further distinguish these models. It involves calculating the critical difference (CD):
$CD=q_\alpha\sqrt{\frac{M(M+1)}{6N}}.$ At a significance level of 0.05, $q_\alpha$ is 2.949. To visualize the differences among seven classifiers, we generate a CD diagram depicted in Figure \ref{fig:CD}. This figure clearly shows that the horizontal line representing the GBFTSVM classifier does not overlap with those of the TWSVM classifier, the EFSVM classifier, the IFTSVM classifier, the GBSVM classifier, and the GBFSVM classifier. This indicates that the GBFTSVM classifier significantly differs from and outperforms other models.

\section{Conclusions}
\label{sec:six}

Inspired by GBC and TWSVM, we have proposed the GBTWSVM classifier for binary classification problems, which utilizes granular-balls as inputs instead of individual samples, and provides a scalable, efficient, and robust data processing method.
Subsequently, we have introduced the GBFTSVM classifier by integrating GBC, PFS, and FTSVM, in which it defines the granular-ball membership and non-membership functions based on the Pythagorean closeness index. Additionally, the GBFTSVM classifier distinguishes the contribution of granular-balls from different regions to classification. Finally, the experimental results on 20 benchmark datasets have demonstrated that the GBFTSVM classifier exhibits excellent results in terms of classification $Acc$, while the GBTWSVM classifier shows superior performance in training time.

In real-world applications, there are numerous instances of nonlinear classification problems and multi-classification challenges. Our forthcoming research will be dedicated to proficient classifiers and algorithms specifically designed to tackle these diverse challenges. 

\section*{Acknowledgements}

This work is supported by the National Key Research and Development Program (No. 2022YFB3104700), the National Natural Science Foundation of China (No. 62076040, 61976158, 62376198), Hunan Provincial Natural Science Foundation of China (No. 2020JJ3034), and the Scientific Research Fund of Hunan Provincial
Education Department (No. 22A0233).

\section*{Appendix: Abbreviations}

\begin{table}[H]\renewcommand{\arraystretch}{1.1}
	\centering
	\label{table:datasets}
	\tabcolsep0.070in
	\begin{tabular}{|c|l|}
		\hline
		\textbf{Abbreviation}   & \textbf{Full Term}\\ \hline
		GC         & Granular Computing\\
		PFS        & Pythagorean Fuzzy Sets\\
		SVM        & Support Vector Machine\\
		GBC        & Granular-Ball Computing\\
		GBRS       & Granular-Ball Rough Set\\
		FSVM       & Fuzzy Support Vector Machine\\
		TWSVM      & Twin Support Vector Machine\\
		GBSVM      & Granular-Ball Support Vector Machine\\
		FTSVM      & Fuzzy Twin	Support Vector Machine\\
		EFSVM      & Entropy-based Fuzzy Support Vector Machine\\
		IFTSVM     & Intuitionistic Fuzzy Twin Support Vector Machine\\
		GBFSVM     & Granular-Ball Fuzzy Support Vector Machine\\
		GBTWSVM    & Granular-Ball Twin Support Vector Machine\\
		GBFTSVM    & Granular-Ball Fuzzy Twin Support Vector Machine\\
		\hline
	\end{tabular}
\end{table}

\end{document}